\documentclass{article} 
\usepackage{iclr2025}

\usepackage{url}
\definecolor{citation}{RGB}{10,110,150}  
\usepackage{hyperref}
\hypersetup{
    colorlinks=true,
    linkcolor=citation,
    anchorcolor=citation,
    citecolor=citation
}
\usepackage{times}
\usepackage{xcolor}
\usepackage{color}
\usepackage{soul}
\usepackage[most]{tcolorbox}
\usepackage{graphicx}
\usepackage{subcaption}
\usepackage{xspace}
\usepackage{algorithm}
\usepackage{amsmath}
\usepackage{amssymb} 
\usepackage{algpseudocode}
\usepackage{booktabs}
\usepackage{multicol}
\usepackage{multirow}
\usepackage{colortbl}
\usepackage{enumitem}
\usepackage{fancyvrb}
\usepackage{mdframed}
\usepackage{threeparttable}
\usepackage{pifont}
\usepackage{wrapfig}
\usepackage{fontawesome5}
\usepackage{soul}

\definecolor{Periwinkle}{HTML}{7977B8}
\definecolor{ProcessBlue}{HTML}{00B0F0}
\definecolor{Cerulean}{HTML}{00A2E3}
\definecolor{OliveGreen}{HTML}{3C8031}
\definecolor{TealBlue}{HTML}{00AEB3}
\definecolor{RoyalPurple}{HTML}{613F99}
\definecolor{RawSienna}{HTML}{974006}
\definecolor{RedOrange}{HTML}{F26035}
\definecolor{PineGreen}{HTML}{008B72}
\definecolor{MidnightBlue}{HTML}{006795}
\definecolor{Aquamarine}{HTML}{00B5BE}
\definecolor{SeaGreen}{HTML}{3FBC9D}
\definecolor{Blue}{HTML}{2D2F92}
\definecolor{NavyBlue}{HTML}{006EB8}
\definecolor{Fuchsia}{HTML}{8C368C}
\definecolor{CadetBlue}{HTML}{74729A}
\definecolor{LimeGreen}{HTML}{8DC73E}
\definecolor{Green}{HTML}{00A64F}
\definecolor{Gray}{gray}{0.5}

\definecolor{darkergreen}{RGB}{0, 180, 0}
\definecolor{darkeryellow}{RGB}{230, 215, 100}

\newcommand{\yes}{\textcolor{darkergreen}{\checkmark}}        
\newcommand{\no }{\textcolor{red}{\ding{55}}}           
\newcommand{\half}{
  \begin{tikzpicture}[baseline=(C.base)]
    \node[inner sep=0pt, darkeryellow] (C) {\checkmark}; 
    \draw[darkeryellow, thick, line cap=round, overlay]
      ([xshift=-0.18em, yshift=0.23em]C.center) -- ([xshift=0.22em, yshift=-0.28em]C.center);
  \end{tikzpicture}%
}

\newcommand{\header}[1]{\smallskip \noindent\textbf{{#1}}\xspace}

\newcommand{\method}{\textsc{Double-Bench}\xspace}

\newcommand\blfootnote[1]{%
  \begingroup
  \renewcommand\thefootnote{}\footnote{#1}%
  \addtocounter{footnote}{-1}%
  \endgroup
}
\title{\includegraphics[height=2em]{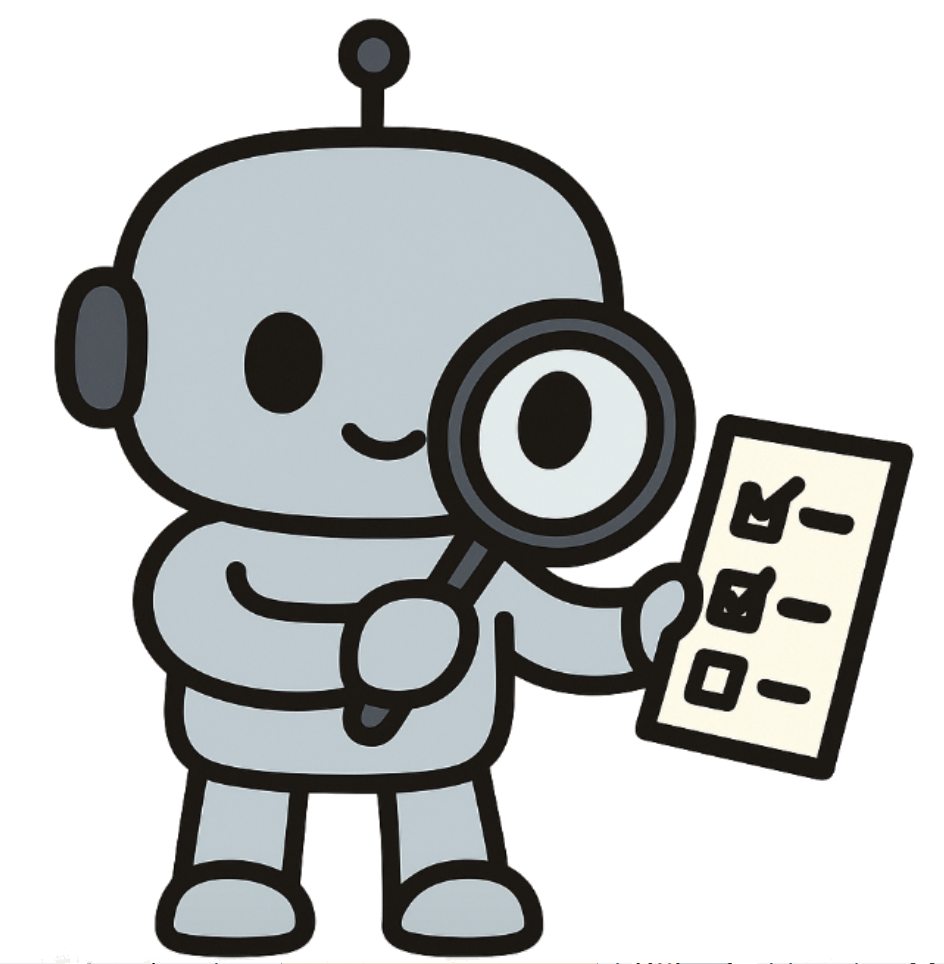}Are We on the Right Way for Assessing Document Retrieval-Augmented Generation?}

\author{
  \hspace{-0.25cm}\textbf{Wenxuan Shen$^{1*}$, Mingjia Wang$^{2*}$, Yaochen Wang$^{2}$, Dongping Chen$^{3\ddagger}$, Junjie Yang$^{1}$,}  \vspace{0.2em} \\
  \hspace{-0.15cm}\textbf{Yao Wan$^{2}$, Weiwei Lin$^{1\dagger}$} \vspace{1em}\\
  \hspace{-0.25cm}$^1$South China University of Technology,\\ \hspace{-0.25cm}$^2$Huazhong University of Science and Technology, \\
  \hspace{-0.25cm}$^3$University of Maryland \\
  }

\iclrfinalcopy 
\begin{document}
\maketitle

\begin{center}
    \vspace{-1.5em}
    \textbf{\url{https://double-bench.github.io}}
    \vspace{1em}
\end{center}

\blfootnote{$^{*}$Equal Contribution, $^{\ddagger}$Project Lead, $^{\dagger}$Corresponding Authr}

\begin{abstract}
Retrieval-Augmented Generation (RAG) systems using Multimodal Large Language Models (MLLMs) show great promise for complex document understanding, yet their development is critically hampered by inadequate evaluation. Current benchmarks often focus on specific part of document RAG system and use synthetic data with incomplete ground truth and evidence labels, therefore failing to reflect real-world bottlenecks and challenges. To overcome these limitations, we introduce \method: a new large-scale, multilingual, and multimodal evaluation system that is able to produce fine-grained assessment to each component within document RAG systems. It comprises 3,276 documents (72,880 pages) and 5,168 \textit{single-} and \textit{multi-hop} queries across 6 languages and 4 document types with streamlined dynamic update support for potential data contamination issues. Queries are grounded in exhaustively scanned evidence pages and verified by human experts to ensure maximum quality and completeness. Our comprehensive experiments across 9 \emph{state-of-the-art} embedding models, 4 MLLMs and 4 \emph{end-to-end} document RAG frameworks demonstrate the gap between text and visual embedding models is narrowing, highlighting the need in building stronger document retrieval models. Our findings also reveal the over-confidence dilemma within current document RAG frameworks that tend to provide answer even without evidence support. We hope our fully open-source \method provide a rigorous foundation for future research in advanced document RAG systems. We plan to retrieve timely corpus and release new benchmarks on an annual basis.

\end{abstract}

\section{Introduction}

Retrieval-Augmented Generation (RAG) \citep{lewis2020retrieval} has emerged as a transformative technique in textual information retrieval, enhancing Large Language Models (LLMs) by retrieving the most relevant content from knowledge bases in response to queries. This approach has driven significant advances in context engineering, particularly for knowledge-intensive NLP tasks. Besides text-only scenarios, vision documents---including scanned files \citep{breci2024novel}, charts \citep{masry2022chartqa}, and slides \citep{tanaka2023slidevqa}---serve as rich information sources that have traditionally required substantial manual effort to examine. These documents are now being efficiently enhanced and handled through multimodal document RAG systems, enabling advanced document understanding~\citep{faysse2024colpali,wang2025vidorag,cho2024m3docrag}. 

\begin{wrapfigure}{r}{0.5\linewidth}
    \centering
    \vspace{-1em}
    \includegraphics[width=1\linewidth]{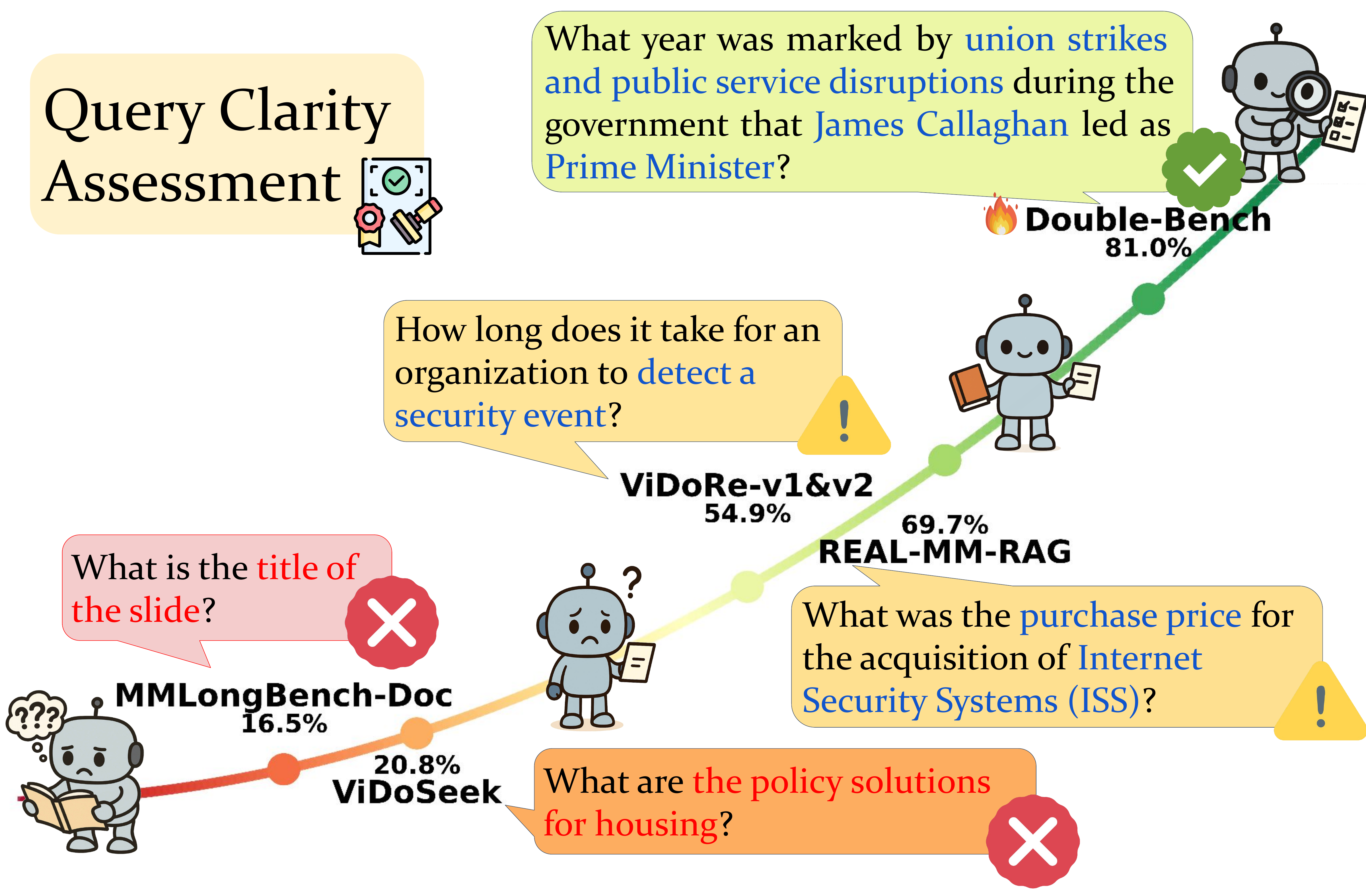}
    \caption{Existing document RAG benchmarks suffer from ambiguous queries that are insufficiently specified to retrieve relevant results, failing to authentically evaluate current document retrieval models and systems.
    }
    \label{fig:pilot study statistics}
    \vspace{-1em}
\end{wrapfigure}

Despite the growing importance of document RAG systems \citep{mortaheb2025rag, yu2024evaluation,mortaheb2025re}, effectively evaluating them in detail presents significant challenges. Existing document RAG evaluation benchmarks \citep{friel2024ragbench} suffer from four critical shortcomings as we identified through pilot experiments shown in Figure \ref{fig:pilot study statistics}: \textbf{(1) Limited evaluation scope}: Current benchmarks only focus on specific parts such as embedding model or VQA model, failing to reveal the bottlenecks of current RAG system in a holistic and comprehensive way. \textbf{(2) Unrealistic prior knowledge assumptions:} Many VQA-style benchmarks \citep{li2024benchmarking,wu2025visual} presuppose that the target page or document is already known, making queries inappropriate for evaluating real-world global retrieval scenarios. \textbf{(3) Ambiguous or non-unique evidence:} Synthetic queries are often crafted from a single page and assume a one-to-one mapping between query and evidence, neglecting cases where multiple pages could be relevant. \textbf{(4) Unlinked multi-hop compositions:} Synthesized multi-hop queries are frequently constructed from loosely connected single-hops, failing to evaluate retrieval models' ability on multi-hop reasoning across documents and modalities.

\begin{figure*}[!t]
    \centering
    \includegraphics[width=1\linewidth]{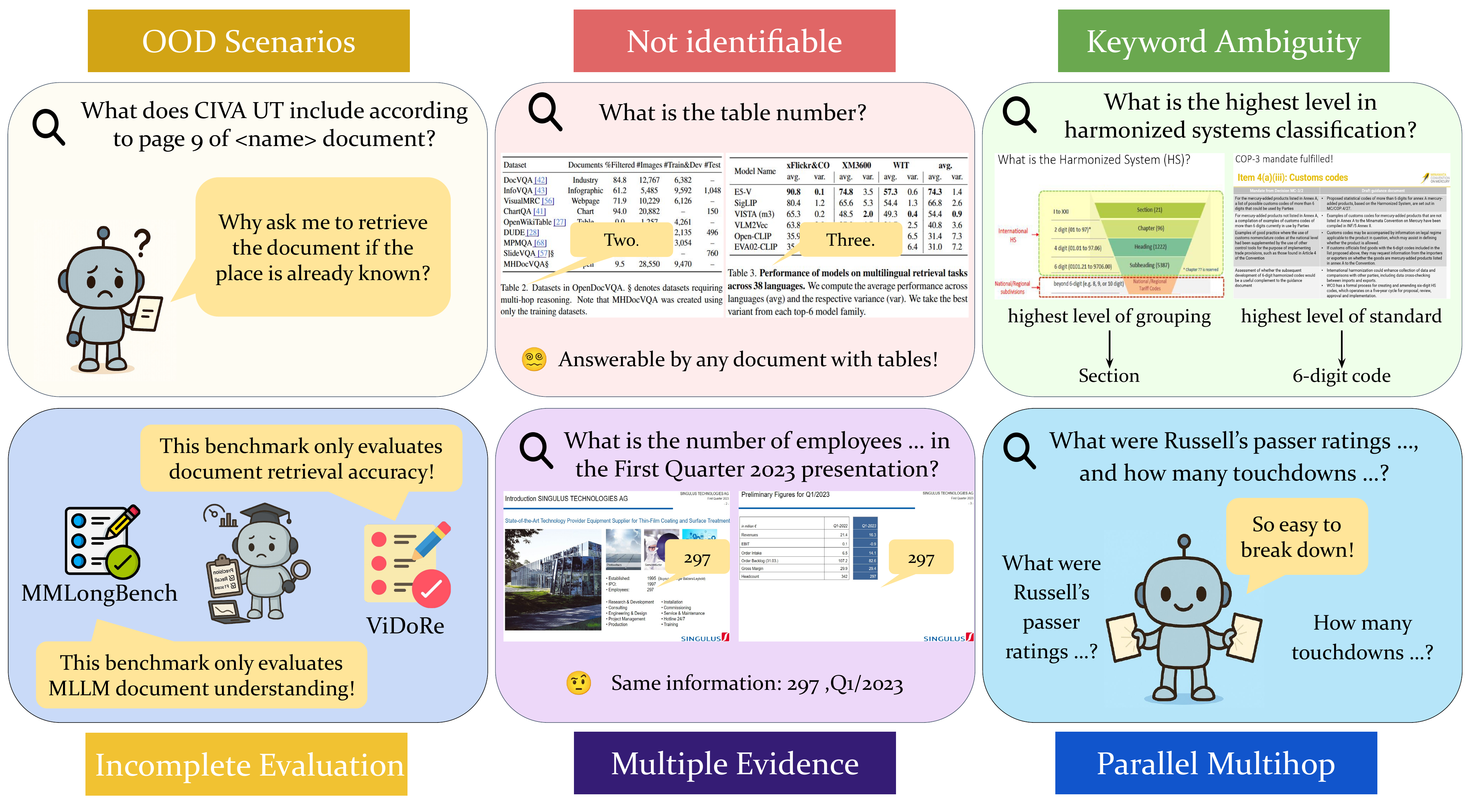}
    \vspace{-1em}
    \caption{Our pilot study reveals critical limitations in existing document RAG system benchmarks, which make them unable to reliably assess current system in a fine-grained and realistic manner.}
    \label{fig:pilot study}
    \vspace{-1em}
\end{figure*}

To overcome these limitations, we introduce \method(\textbf{DO}cument \textbf{U}nified \textbf{B}road-coverage \textbf{L}ogical hops \textbf{E}valuation Benchmark), consisting of 5,168 human-validated single- and multi-hop queries and 3,276 documents in 6 languages and 4 types of document data. First, we assemble and preprocess a large diverse document corpus spanning PDFs, scanned documents, slides and web pages by a two-stage filtering and modality decomposition. Next, we synthesize and rigorously label both single- and multi-hop queries \citep{zhang2025levelrag, tang2024multihop}, using a combination of iterative clarity-oriented refinement, knowledge-graph-guided composition and exhaustive evidence search. Finally, expert annotators review \citep{chiang2024chatbot, chen2024mllm} and correct machine-assigned evidence to ensure high-precision ground truth for large-scale realistic multimodal retrieval evaluation. As shown in Figure \ref{fig:pilot study}, \method contains high quality queries with low ambiguity. To avoid potential data contamination issues, \method is also designed to support streamlined dynamic updates with minimal human intervention.

Based on \method, we conduct extensive experiments across 9 \emph{state-of-the-art} textual, visual and multimodal embedding models, 4 MLLMs and 4 advanced document RAG frameworks. Our findings demonstrate that text embedding models are narrowing the gap with visual embedding models, and \texttt{Colqwen2.5-3b} achieve a strong performance with an 0.795 averaged hit@5 score. Regarding different languages, retrieval models perform generally better on high-resource language than low-resource like Arabic and French. Regarding different document types, clean and structured documents, such as PDFs and HTML pages, are generally easier for models to inspect. Moreover, MLLMs' low accuracy across both single- and multi-hop queries demonstrates the inherent challenges in multimodal long document understanding. Multi-hop queries prove particularly challenging for current document RAG frameworks, achieving only 0.655 accuracy even when ground truth pages are directly provided.

In summary, our contributions are three-fold:
\begin{itemize}[leftmargin=*,itemsep=0pt]
    \item We diagnose several major limitations in existing document RAG evaluation, including incomplete scope, unrealistic prior knowledge assumptions, ambiguous or non-unique evidence, and non-grounded multi-hop query design. 
    \item We introduce \method, the first-of-its-kind live evaluation system for multilingual and multimodal document RAG system, featuring a diverse document corpus, fine-grained page decomposition, and high-quality \textit{single-} and \textit{multi-hop} QA pairs with manually labeled evidence.
    \item Our experiments across several \emph{state-of-the-art} embedding models, MLLMs, and document RAG system uncover critical limitations in current RAG frameworks, providing insights and findings for the research community.
\end{itemize}

\section{Limitations of Existing Document RAG System Evaluation: A Pilot Study}

\label{section-pilot study}
\subsection{Task Formulation} 
\label{section-task formulation}
Let $C$ be a large corpora consisting of documents $\{ d_1, d_2, ..., d_n\}$. Each document $d_i$ is stored by page images $\{ p_1^i, p_2^i, ..., p_m^i\}$. Given a query $Q$, the objective is to retrieve top-$k$ possible evidence pages $E_r$ from the entire corpora to formulate the answer $A$. For single-hop queries $Q_{s}$, answer $A$ can be found if one or more evidence pages from the evidence set $E_q$ is successfully retrieved. For multi-hop queries $Q_{m}$, the requirement extends to having one or more evidence page for every evidence set $E_{q,j}$ of each query hop $j$. These conditions are formated in Equation~\eqref{eq1} and \eqref{eq2}.

\label{Formula: Evaluation}
\begin{equation}\label{eq1}
(E_r \cap E_q \neq \varnothing) \implies \text{Enable}(A|Q_{s}, E_r)
\end{equation}

\begin{equation}\label{eq2}
\resizebox{0.53\linewidth}{!}{%
  $\displaystyle \left( \bigwedge_{j=1}^{k} (E_r \cap E_{q,j} \neq \varnothing) \right) \implies \text{Enable}(A|Q_{m},E_r)$%
}
\end{equation}

\subsection{Four Major Overlooked Issues for Document RAG Evaluation}

Though practical document RAG scenarios typically involve queries that clearly state an informational need, the core utility of RAG is demonstrated when users do not possess specific prior knowledge about individual documents, such as their titles, filenames, or the precise location of some document component, even if users have general familiarity with the topics within the collection.

Therefore, we start by investigating whether existing benchmarks are fully appropriate for evaluating real-world document RAG scenarios. By screening existing benchmarks with concrete rules, we illustrate these limitations in Figure~\ref{fig:pilot study}.

\begin{table*}[!t]
\centering
\setlength{\tabcolsep}{6pt}
\caption{Comparison between existing multimodal document related benchmarks and the proposed \method, where each symbol represents:\hspace{0.1em} \faFilePdf \hspace{0.05em} PDFs; \quad \faFileSignature \hspace{0.05em} Scanned documents; \quad \faChalkboardTeacher \hspace{0.05em} Slides; \quad \faGlobe \hspace{0.05em} HTML pages. Half-tick denotes dependent on the specific benchmark component or insufficient evaluation. \textbf{\textit{GT}}: Ground Truth evidence lables. \textbf{\textit{M.H.}}: Multi-Hop. \textbf{\textit{Lang.}}: Suppoorted language number. \textbf{\textit{Dyna.}}: Support dynamic benchmark update when datapoint are contamintaed.}

\label{comparison table - table 1}

\resizebox{\textwidth}{!}{
\begin{tabular}{l|ccc|ccc|cc|ccc|ccp{0.4em}p{0.4em}p{0.4em}p{0.4em}}
\toprule[1.5pt]
\multirow{2}{*}{\textbf{Benchmarks}} & \multicolumn{3}{c|}{\textbf{Size}}& \multicolumn{3}{c|}{\textbf{Queries}} & \multicolumn{2}{c|}{\textbf{Labels}} & \multicolumn{3}{c|}{\textbf{Evaluation Target}} & \multicolumn{6}{c}{\textbf{Document}} \\
& Doc & Avg. \#Pages & Query & Clarity & i.i.d. & M.H. & GT & M.H. Chain & Embed Model& MLLMs & RAG System & Lang. & Dyna. & \multicolumn{4}{c}{Type} \\
\midrule
DocVQA & 6,071 & 1.0 & 50,000 & \no & \no & \no & \no &  - & \no & \yes  & \no & 1 & \no & \faFilePdf & \faFileSignature & \phantom{\faChalkboardTeacher} & \phantom{\faGlobe} \\
MMLongbench-Doc & 135 & 47.5 & 1,082 & \no & \half & \yes & \no & \no & \no & \yes & \no & 1 & \no & \faFilePdf & \phantom{\faFileSignature} & \phantom{\faChalkboardTeacher} & \phantom{\faGlobe} \\
MMDocIR & 6,818 & 65.1 & 73,843& \no & \no & \yes & \no & \no  & \yes & \no & \no & 1 & \no& \faFilePdf & \phantom{\faFileSignature} & \phantom{\faChalkboardTeacher} & \phantom{\faGlobe} \\
UDA-QA & 2,965 & 46.3 & 29,590 & \no & \yes & \half & \no & \no & \half & \no & \no & 1 & \no& \faFilePdf & \phantom{\faFileSignature} & \phantom{\faChalkboardTeacher} & \faGlobe \\
ViDoRe v1 & 5,000 & 1.0 & 500 & \half & \yes & \no & \half & -  & \yes & \no & \no & 2 & \no & \faFilePdf & \faFileSignature & \faChalkboardTeacher & \phantom{\faGlobe} \\
ViDoRe v2 & 65 & 48.6 & 913 & \yes & \yes & \no & \yes & -  & \yes & \no & \no & 2 & \no&  \faFilePdf & \phantom{\faFileSignature} & \faChalkboardTeacher & \phantom{\faGlobe} \\
ViDoSeek & 1,142 & 18.4 & 1,142 & \yes & \no & \yes & \no & \no  & \yes & \no & \half & 1 & \no& \phantom{\faFilePdf} & \phantom{\faFileSignature} & \faChalkboardTeacher & \phantom{\faGlobe} \\
REAL-MM-RAG & 163 & 49.1 & 4,553 & \yes & \yes & \no & \yes & -  & \yes & \no & \no & 1 & \no& \phantom{\faFilePdf} & \phantom{\faFileSignature} & \faChalkboardTeacher & \phantom{\faGlobe} \\
\midrule
\textbf{\method} & 3,276 & 22.3 & 5,168 & \yes & \yes & \yes & \yes & \yes  & \yes  & \yes & \yes  & \textbf{6} & \yes & \faFilePdf & \faFileSignature & \faChalkboardTeacher & \faGlobe \\
\bottomrule[1.5pt]
\end{tabular}
}
\vspace{-1em}
\end{table*}

\header{Current benchmarks fail to comprehensively evaluate document RAG systems with fine-grained breakdown.} As shown in Table \ref{comparison table - table 1}, current document-related benchmarks usually focus on embedding-based retrieval models \citep{mace2025vidore} or response models \citep{ma2024mmlongbench}, which are only components within document RAG systems, failing to provide a comprehensive assessment. This fragmented evaluation obscures critical interaction effects between retrieval and generation components that often determine real-world system performance.

\header{Benchmark design issues with prior knowledge assumptions.} VQA benchmarks, such as MMLongbench-Doc \citep{ma2024mmlongbench}, DocVQA \citep{mathew2021docvqa}, and MMDocIR \citep{dong2025mmdocir}, are inherently designed with a given page or document as prior knowledge, rendering their queries ambiguous for identifying the ground-truth page within a global document corpus. Manually inspected benchmarks designated for RAG, such as ViDoSeek \citep{wang2025vidorag} and MRAMG-Bench \citep{yu2025mramg}, have significant gains in query information. However, such benchmarks tend to insert the exact name or page of the ground document, failing to align with intended applications where users do not have any specific prior knowledge over individual documents. Such queries create gaps between evaluation and real use.

\header{Queries with multiple interpretations and scattered evidence.} Most benchmarks construct queries by selecting a ground truth page beforehand, and assume the evidence used is unique \citep{chen2024benchmarking, tang2024multihop}. This generally holds true when the corpus is small, such as individual benchmarks in ViDoRe \citep{faysse2024colpali}, but the problem becomes uncontrolled when the corpus scales up. Some queries may also have unexpected multiple interpretations given different content in the same document, which further undermines the unique assumption.

\header{The linearity in multi-hop query synthesis is overlooked.}
The inclusion of trivial multi-hop queries in evaluations may overstate the reasoning capabilities of RAG frameworks, inflating perceived performance without accurately assessing genuine multi-step reasoning. These queries are essentially simple linkings of independent parts, do not necessitate complex reasoning to deconstruct and can be processed in parallel \citep{hui2024uda}. 

\section{\method: The Benchmark}
\begin{figure*}[htb]
\centering
\includegraphics[width=1\linewidth]{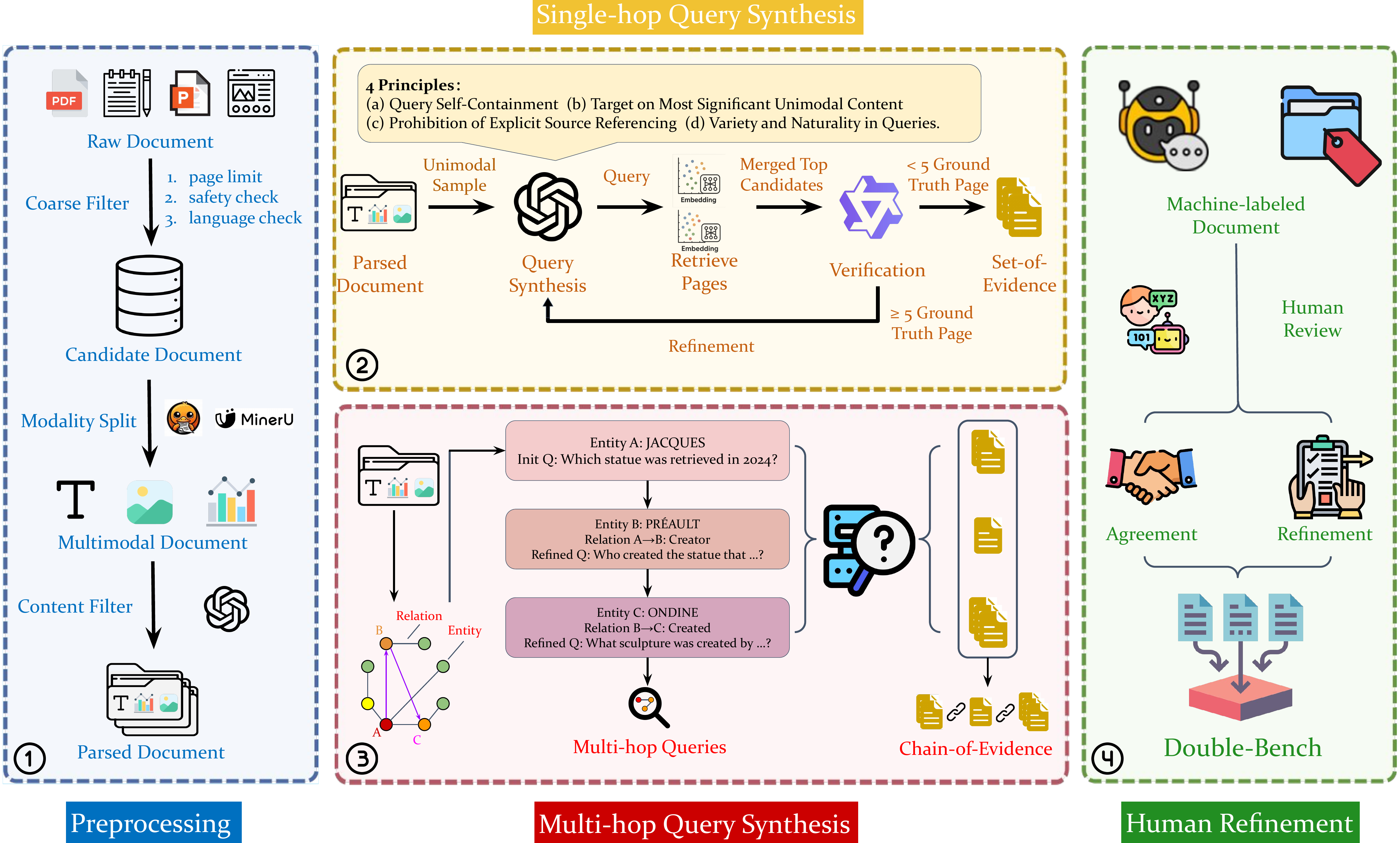}
\vspace{-1.5em}

\caption{Overview of the \method construction pipeline. The preprocessing stage filters the collected corpora and splits the content by modality. To alleviate identifed problems explained in Figure~\ref{fig:pilot study}, an iterative clarity-oriented refinement pipeline is introduced for single-hop query generation, while knowledge graphs are additionally constructed to assist multi-hop query generation. All document pages are thoroughly checked by annotators to produce list-of-evidence and set-of-evidence labels.}

\vspace{-1em}
\label{fig:pipeline}
\end{figure*}

To address existing limitations, we introduce \method, a benchmark with manually verified multi-modal, multi-lingual content and an automatic benchmark construction suite via a three-stage pipeline shown in Figure \ref{fig:pipeline}. Detailed benchmark statistics can be found in Figure \ref{fig:distribution} and Appendix.  We also provide extensive metadata, such as queried modality, language, evidence chains/lists and parsed page chunks to advance document RAG research community.

\subsection{Metadata Collection and Preprocessing}
This section details the preprocessing steps applied to the raw document corpus: \textbf{\textit{(1)}} Large Corpus Collection, \textbf{\textit{(2)}} Two-stage Filtering, and \textbf{\textit{(3)}} Modality Split.

\begin{wrapfigure}{r}{0.4\linewidth}
    \centering
    \vspace{-1em}
    \includegraphics[width=\linewidth]{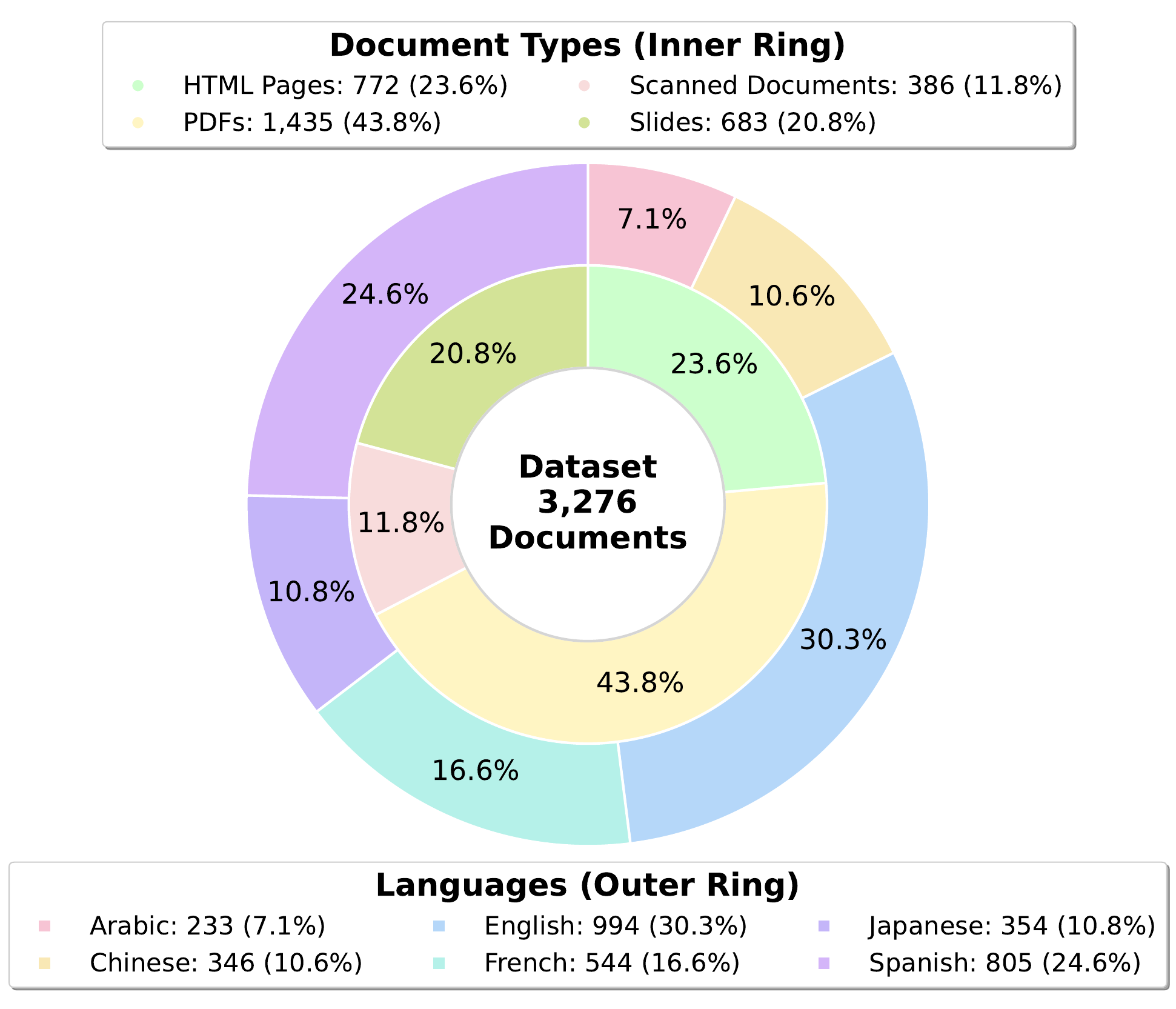}
    \caption{Statistics of the \method dataset. See Appendix for more details.
    }
    \label{fig:distribution}
    \vspace{-1em}
\end{wrapfigure}

\header{Metadata Collection.}
To ensure comprehensive evaluation, we collect a diverse range of document types and languages. The initial database comprises four popular types of documents collected from various sources. As shown in Figure \ref{fig:distribution}, we include high-quality PDF files, scanned documents, slides, and HTML pages to ensure the diversity coverage of our raw data:

\begin{itemize}[leftmargin=*,itemsep=0pt]
\item \textbf{PDFs:} Includes high-quality PDFs from DocVQA \citep{mathew2021docvqa}, MMLongBench-Doc \citep{ma2024mmlongbench}, and Arxiv papers published by reputable organizations after 2023, all known for rich context and multimodal content. 
\item \textbf{Scanned Documents:} Printed or handwritten documents sampled from DocVQA \citep{mathew2021docvqa}, MMLongBench-Doc \citep{ma2024mmlongbench}, and the CommonCrawl corpus. Scanned content presents greater challenges for models due to defects introduced by the document creation and scanning process, \emph{e.g.}, variability in font styles, layout irregularities, background noise.
\item \textbf{Slides:} A subset from SlideVQA \citep{tanaka2023slidevqa}, augmented with multilingual slides from the Commoncrawl corpus. Collected slides are scanned with Qwen2.5-VL \citep{bai2025qwen2}, and only those with at least 30\% multimodal pages remain.
\item \textbf{HTML Pages:} 600 Wikipedia entries, randomly crawled across various topics, with equal non-overlapping samples per language.
\end{itemize}

\header{Filtering and Preprocessing.}
Our preprocessing pipeline begins with a coarse-grained, rule-based filter \citep{chen2024mllmasajudge,pu2025judge} to select documents that meet basic structural and language requirements. GPT-4o \citep{openai2024gpt4o} reads the first three pages of every document to determine the primary language. Only documents with 10 to 50 pages and a primary language listed in Figure~\ref{fig:distribution} are retained. Following this initial selection, each document page undergoes a modality split, where it is parsed and decomposed into its constituent text, table, and figure components using tools like Docling \citep{livathinos2025docling} and MinerU \citep{wang2024mineru}. Finally, we apply a fine-grained content filter where parsed chunks are reviewed with their adjacent context to ensure semantic coherence, filtering out any content that is irrelevant or lacks meaningful information for query generation.




\subsection{Single-hop Query Synthesis}

\header{Principles.}
Single-hop VQA queries often lack enough detail for precise document retrieval. We enhance them by adding supportive descriptions, making queries self-contained, focused on key unimodal information, and diverse in type including factual and analytical. This produces robust queries for evaluating single-hop retrieval.

\header{Synthesis Process.}
The Single-hop query synthesis process goes through an iterative refinement process to ensure clarity. Provided a parsed page component obtained by preprocessing, we leverage \texttt{GPT-4o} to formulate an initial query based on the following four principles: (1) Query self-containment; (2) Target on most significant unimodal content; (3) Prohibition of explicit source eeferencing; (4) Variety and naturality in queries. This initial query is then validated against the corpus. Two high-performance embedding models of different modalities, \texttt{colqwen} and \texttt{qwen3-embedding}, are used to retrieve the top-10 candidate pages each. Subsequently, \texttt{Qwen2.5-VL-32B} processes the merged candidates seperately to identify all ground truth pages containing a direct answer. If more than five ground truth pages are found, the model is prompted to refine the query by incorporating a distinguishing detail extracted from one of the identified ground truth pages. This validation and refinement loop continues until the query yields five or fewer ground truth pages, ensuring the overall difficulty.



\begin{table*}[!tb]
\centering
\caption{Overview of our dataset statistics, grouped by document type.}
\label{tab:dataset summary}
\setlength{\tabcolsep}{5pt} 
\resizebox{\textwidth}{!}{
\begin{tabular}{l|c c|c c c|c c c} 
\toprule[1.5pt]
\textbf{Document Type} & \textbf{Doc Count} & \textbf{Avg. Length} & \textbf{1-hop Qs} & \textbf{2-hop Qs} & \textbf{3-hop Qs} & \textbf{Text Modality} & \textbf{Table Modality} & \textbf{Figure Modality} \\
\midrule
PDFs & 1435 & 22.28 & 1109 & 188 & 900 & 1453 & 915 & 329 \\
Scanned Doc & 386 & 21.49 & 136 & 49 & 229 & 271 & 236 & 36 \\
Slides & 683 & 21.94 & 299 & 86 & 261 & 371 & 222 & 139 \\
HTML Pages & 772 & 22.83 & 956 & 256 & 693 & 1158 & 428 & 115 \\
\midrule
\textbf{Total} & \textbf{3276} & \textbf{22.25} & \textbf{2500} & \textbf{579} & \textbf{2083} & \textbf{3253} & \textbf{1801} & \textbf{619} \\
\bottomrule[1.5pt]
\end{tabular}
}
\end{table*}

\subsection{Multi-hop Query Synthesis}

\header{Principles.} While multi-hop queries benefit from information across hops, which mitigates the lack of information problem, their direct generation by LLMs is challenging, even with techniques like Chain-of-Thought or inference scaling. Problems in current synthetic multi-hop benchmarks, such as trivial connections and non-realistic intents, have undermined accurate evaluations of RAG frameworks. Our multi-hop query synthesis pipeline addresses this by using knowledge graphs and intent-driven walks. This approach simplifies sub-query linking by replacing key entities with new sub-queries, forming linearly combined queries.

\header{Synthesis Process.}
Our multi-hop query synthesis process uses a knowledge graph-based approach to generate complex reasoning chains. We construct knowledge graphs for each document using LightRAG \citep{guo2024lightrag}, ensuring extracted relationships uniquely identify target entities to prevent sub-query ambiguity. An LLM agent selects an initial node from high-degree entities, infers query intent based on the node and document summary, then performs a guided graph walk by iteratively selecting neighbors that best align with the inferred intent. The final multi-hop question is built iteratively along this path: for each step, a sub-query is generated and nested within the cumulative query by identifying the queried entity, replacing it with the sub-query, and rearranging for natural flow. This transforms a simple entity-relation path into a grammatically natural, complex question requiring sequential reasoning to answer.

\subsection{Post-Processing}
\header{Query Quality Inspection.}
The query drafting module strives to generate high-quality queries, but a quality inspection is required to ensure all criteria are met. For single-hop queries, the checklist reviews all generation requirements. For multi-hop queries, we create a separate checklist, assessing: (1) Final question quality (clarity, specificity, no explicit final answer); (2) Logical necessity and correctness of intermediate reasoning steps; (3) Uniqueness of step answers and rigor of relations; and (4) Significance and relevance of the overall query. Queries failing any criteria are discarded.

\header{Evidence Labeling.}
For each query, we locate all ground truth by thoroughly searching each page within the document. Pages are marked as evidence only if they directly provide or lead to the answer. We provide set-of-evidence labels for single-hop queries---a set of all evidence pages. Chain-of-evidence labels, which distinguishes the set-of-evidence labels for every hop, is provided for multi-hop queries.

\header{Human Refinement.}
To improve benchmark quality and ensure accuracy, we further conduct human refinement. Although automated evidence labeling is sufficiently accurate, human annotators reviewed and adjusted 8\% of labels with discrepancies. With 92\% agreement, this step ensures precise ground truth data, enhancing \method’s reliability.

\begin{table*}[!tb]
\centering
\setlength{\tabcolsep}{8pt}
\caption{Retrieval accuracy of \emph{state-of-the-art} text and multimodal embedding models across query types, showing performance degradation as reasoning complexity increases.}
\label{tab:Emb_Model_Results}
\resizebox{1\textwidth}{!}{
\begin{tabular}{l|ccc|ccc|ccc|ccc}
\toprule[1.5pt]
\multirow{2}{*}{\textbf{Model}} & \multicolumn{3}{c|}{\textbf{Average}} & \multicolumn{3}{c|}{\textbf{Single Hop}} & \multicolumn{3}{c|}{\textbf{2-Hop}} & \multicolumn{3}{c}{\textbf{3-Hop}} \\ \cmidrule{2-13}
 & hit@1 & hit@3 & \multicolumn{1}{c|}{hit@5} & hit@1 & hit@3 & \multicolumn{1}{c|}{hit@5} & hit@1 & hit@3 & \multicolumn{1}{c|}{hit@5} & hit@1 & hit@3 & \multicolumn{1}{c}{hit@5} \\
\midrule
\multicolumn{13}{c}{\textcolor{violet!80!white}{\cellcolor{gray!10!white} \textbf{\textit{Text Embedding Models}}}}\\
\midrule
\multicolumn{1}{l|}{Qwen3-Embedding-4B} & \textbf{0.489} & \textbf{0.699} & \multicolumn{1}{c|}{\textbf{0.776}} & \textbf{0.726} & \textbf{0.852} & \multicolumn{1}{c|}{\textbf{0.886}} & 0.314 & 0.598 & \multicolumn{1}{c|}{0.663} & 0.235 & \textbf{0.531} & \multicolumn{1}{c}{\textbf{0.668}} \\
\multicolumn{1}{l|}{NV-Embed-v2} & 0.443 & 0.650 & \multicolumn{1}{c|}{0.724} & 0.626 & 0.756 & \multicolumn{1}{c|}{0.796} & \textbf{0.333} & \textbf{0.604} & \multicolumn{1}{c|}{\textbf{0.689}} & \textbf{0.240} & 0.526 & \multicolumn{1}{c}{0.641} \\
\multicolumn{1}{l|}{gte-Qwen2-7B-instruct} & 0.404 & 0.611 & \multicolumn{1}{c|}{0.697} & 0.585 & 0.749 & \multicolumn{1}{c|}{0.804} & 0.288 & 0.503 & \multicolumn{1}{c|}{0.603} & 0.205 & 0.466 & \multicolumn{1}{c}{0.588} \\
\multicolumn{1}{l|}{bge-m3} & 0.355 & 0.525 & \multicolumn{1}{c|}{0.591} & 0.527 & 0.648 & \multicolumn{1}{c|}{0.695} & 0.180 & 0.366 & \multicolumn{1}{c|}{0.428} & 0.182 & 0.412 & \multicolumn{1}{c}{0.502} \\
\midrule 
\multicolumn{13}{c}{\textcolor{violet!80!white}{\cellcolor{gray!10!white} \textbf{\textit{Visual \& Multimodal Embedding Models}}}}\\
\midrule
\multicolumn{1}{l|}{colqwen2.5-3b-multilingual} & \textbf{0.533} & \textbf{0.727} & \multicolumn{1}{c|}{\textbf{0.795}} & \textbf{0.778} & \textbf{0.865} & \multicolumn{1}{c|}{\textbf{0.895}} & \textbf{0.326} & \textbf{0.622} & \multicolumn{1}{c|}{\textbf{0.693}} & \textbf{0.277} & \textbf{0.579} & \multicolumn{1}{c}{\textbf{0.696}} \\
\multicolumn{1}{l|}{vdr-2b-multi} & 0.463 & 0.648 & \multicolumn{1}{c|}{0.725} & 0.688 & 0.813 & \multicolumn{1}{c|}{0.847} & 0.283 & 0.491 & \multicolumn{1}{c|}{0.589} & 0.225 & 0.482 & \multicolumn{1}{c}{0.606} \\
\multicolumn{1}{l|}{jina-embeddings-v4} & 0.451 & 0.641 & \multicolumn{1}{c|}{0.720} & 0.671 & 0.804 & \multicolumn{1}{c|}{0.844} & 0.264 & 0.468 & \multicolumn{1}{c|}{0.570} & 0.222 & 0.479 & \multicolumn{1}{c}{0.603} \\
\multicolumn{1}{l|}{gme-Qwen2-VL-7B-Instruct} & 0.428 & 0.614 & \multicolumn{1}{c|}{0.697} & 0.638 & 0.775 & \multicolumn{1}{c|}{0.822} & 0.249 & 0.472 & \multicolumn{1}{c|}{0.579} & 0.208 & 0.449 & \multicolumn{1}{c}{0.570} \\
\multicolumn{1}{l|}{colpali-v1.3} & 0.403 & 0.571 & \multicolumn{1}{c|}{0.646} & 0.584 & 0.679 & \multicolumn{1}{c|}{0.717} & 0.230 & 0.440 & \multicolumn{1}{c|}{0.525} & 0.220 & 0.469 & \multicolumn{1}{c}{0.588} \\
\bottomrule[1.5pt]
\end{tabular}
}
\end{table*}
\section{Experiments}
\header{Evaluation Protocol.}
Following task formulation setting, we define the hit rate for retrieval accuracy evaluation of single- and multi-hop queries. The accuracy of the final answer is evaluated using LLM-as-a-judge \citep{zheng2023judging}. \texttt{GPT-4o} rates the correctness the generated answer compared to the ground truth answer on a scale of 0 to 10. Answers with a score not lower than 7 count as correct, not higher than 3 count as incorrect, others count as partially correct.

\header{Evaluated Models and Frameworks.}
We evaluate 4 competitive text embedding models, namely \texttt{bge-m3} \citep{chen2024bge}, \texttt{gte-Qwen2} \citep{li2023towards}, \texttt{NV-Embed-v2} \citep{lee2024nv}, \texttt{Qwen3-Embedding} \citep{zhang2025qwen3}, 5 competitive open-source document page embedding models, namely \texttt{colpali} \citep{faysse2024colpali}, \texttt{colqwen} \citep{faysse2024colpali}, \texttt{gme} \citep{zhang2024gme}, \texttt{vdr-2b} \citep{vdr-2b-multi-v1-hf}, \texttt{jina-embeddings-v4} \citep{gunther2025jina}, and 3 advanced document RAG frameworks, namely \texttt{M3DocRAG} \citep{cho2024m3docrag}, \texttt{MDocAgent} \citep{han2025mdocagent}, VidoRAG \citep{wang2025vidorag}. To understand how each part of RAG framework affects answer accuracy, we add a reference framework, \texttt{Colqwen-gen}, by directly pairing the strongest embedding model \texttt{Colqwen} with \texttt{GPT-4o}. Baseline results are reported with \texttt{GPT-4o} directly answering the queries without RAG. The Oracle setting estimates the upper-bound performance of RAG frameworks. In this setting, we provide the parsed contents of all the ground truth pages to the MLLM, together with a prompt that instructs the MLLM to first extract relevant information and think before providing the final answer. 

\header{Experiment Setups.}
All experiments utilize 8$\times$A100 server. For embedding model and framework evaluation, all 2500 single-hop and 2668 multi-hop queries are assessed using the entire corpus for retrieval. Text-based embedding models use parsed document chunks, with graphs and figures converted to descriptive captions (generated by \texttt{Qwen2.5-VL-32b-Instruct}) before merging text. In RAG framework evaluation, the same page text is used by \texttt{ViDoRAG} and \texttt{MDocAgent}. \texttt{ViDoRAG} can invoke each component up to twice before final answer generation. All frameworks retrieve 5 most relevant pages to answer the query.

\subsection{Main Results}

\header{Document-specified embedding model outperform general ones, and gap between text and image embedding models is narrowing.} 
\method provides a clear divergence in the retrieval performance of various embedding models, as detailed in Table \ref{tab:Emb_Model_Results}. The model rankings within \method align well with popular text embedding leaderboards MTEB \citep{muennighoff2022mteb} and document retrieval benchmark ViDoRe v2 \citep{mace2025vidore}, demonstrating the robustness of our benchmark.

\begin{table*}[!t]
\centering
\setlength{\tabcolsep}{2pt} 
\caption{Performance of RAG Frameworks. \texttt{Colqwen-gen} achieves comparable performance with \texttt{MDocAgent}, the best among evaluated frameworks. This observation highlights the need for more advanced retrieval stage frameworks.}
\label{tab:RAG_Framework_Results}
\resizebox{\textwidth}{!}{
\begin{tabular}{l|c|ccc|c|ccc|c|ccc|c|ccc}
\toprule[1.5pt]
\multirow{4}{*}{\textbf{Framework}} & \multicolumn{4}{c|}{\textbf{Average}} & \multicolumn{4}{c|}{\textbf{Single Hop}} & \multicolumn{4}{c|}{\textbf{2-Hop}} & \multicolumn{4}{c}{\textbf{3-Hop}}\\ \cmidrule{2-17}
 & \multicolumn{1}{c|}{\textbf{Retrieval}} & \multicolumn{3}{c|}{\textbf{Answer}} & \multicolumn{1}{c|}{\textbf{Retrieval}} & \multicolumn{3}{c|}{\textbf{Answer}} & \multicolumn{1}{c|}{\textbf{Retrieval}} & \multicolumn{3}{c|}{\textbf{Answer}} & \multicolumn{1}{c|}{\textbf{Retrieval}} & \multicolumn{3}{c}{\textbf{Answer}} \\ \cmidrule{2-17}
 & \multicolumn{1}{c|}{hit@5} & \yes & \half & \multicolumn{1}{c|}{\no} &  \multicolumn{1}{c|}{hit@5} & \yes & \half & \no &  \multicolumn{1}{c|}{hit@5} & \yes & \half & \no &  \multicolumn{1}{c|}{hit@5} & \yes & \half & \no \\ \midrule
\multicolumn{1}{l|}{MDocAgent \citep{han2025mdocagent}} & \multicolumn{1}{c|}{0.688} & \textbf{0.645} & 0.126 & \multicolumn{1}{c|}{0.229} & \multicolumn{1}{c|}{0.830} & \textbf{0.757} & 0.132 & \multicolumn{1}{c|}{0.111} & \multicolumn{1}{c|}{0.572} & \textbf{0.567} & 0.065 & \multicolumn{1}{c|}{0.367} & \multicolumn{1}{c|}{0.549} & 0.532 & 0.135 & 0.332 \\
\multicolumn{1}{l|}{ViDoRAG \citep{wang2025vidorag}} & \multicolumn{1}{c|}{0.682} & 0.536 & 0.138 & \multicolumn{1}{c|}{0.326} & \multicolumn{1}{c|}{0.822} &  0.623 & 0.144 &  \multicolumn{1}{c|}{0.233} & 0.539 & 0.457 & 0.112 & 0.431 & 0.544 & 0.447 & 0.137 & 0.416 \\
\multicolumn{1}{l|}{M3DOCRAG \citep{cho2024m3docrag}} & \multicolumn{1}{c|}{0.608} & 0.451 & 0.121 & \multicolumn{1}{c|}{0.428} & \multicolumn{1}{c|}{0.709} & 0.538 & 0.138 & \multicolumn{1}{c|}{0.324} & \multicolumn{1}{c|}{0.490} & 0.330 & 0.088 & \multicolumn{1}{c|}{0.582} & \multicolumn{1}{c|}{0.519} & 0.382 & 0.110 & 0.508 \\ 
\multicolumn{1}{l|}{Colqwen-gen \citep{faysse2024colpali}} & \multicolumn{1}{c|}{\textbf{0.795}} & 0.604 & 0.135 & \multicolumn{1}{c|}{0.261} & \multicolumn{1}{c|}{\textbf{0.895}} & 0.676 & 0.160 & \multicolumn{1}{c|}{0.164} & \multicolumn{1}{c|}{\textbf{0.693}} & 0.462 & 0.143 & \multicolumn{1}{c|}{0.395} & \multicolumn{1}{c|}{\textbf{0.696}} & \textbf{0.554} & 0.100 & 0.346 \\

\bottomrule[1.5pt]
\end{tabular}
}
\end{table*}

\texttt{ColQwen2.5-3B} significantly outperforms general multimodal embedding models like \texttt{jina-embeddings-v4} and \texttt{GME}, achieving a 9\% higher average hit rate and demonstrating strong potential in document retrieval. Other multimodal embedding models show limited capability, even underperforming compared to the purely textual embedding model \texttt{Qwen3-Embedding}. We attribute this to recent advancements of text embedding community's sophisticated training techniques, including complex multi-stage training, dedicated hard negative sampling, and large-scale high-quality data synthesis. These techniques are difficult to transfer to visual embedding models due to training costs, limited text-and-image data, and model structural constraints. Although visual embedding models have inherent advantages for visual content retrieval, the semantic complexity of document RAG tasks negates this advantage. The critical influence of both visual observation and textual understanding abilities incentive combined strategies such as interleaved embedding models and advanced multimodal understanding pipelines.

\begin{wraptable}{r}{0.45\linewidth}
\centering
\vspace{-1em}
\caption{Evaluation of MLLMs' long document understanding capability. \textit{Oracle} means directly providing evidnece page. The stark contrast between \emph{w.o.} RAG and Oracle settings of all MLLMs reflect the high quality and low-contaminated of our \method.}
\label{tab:LMLLMs_Results}
\setlength{\tabcolsep}{3pt} 
\resizebox{0.47\textwidth}{!}{ 
\begin{tabular}{l|ccc|ccc}
\toprule[1.5pt]
\multirow{2}{*}{\textbf{Model \& Setting}} & \multicolumn{3}{c|}{\textbf{Single Hop}} & \multicolumn{3}{c}{\textbf{Multi Hop}} \\
& \multicolumn{1}{c}{\yes} & \multicolumn{1}{c}{\half} & \multicolumn{1}{c|}{\no} & \multicolumn{1}{c}{\yes} & \multicolumn{1}{c}{\half} & \multicolumn{1}{c}{\no} \\
\midrule
\multicolumn{7}{c}{\textcolor{violet!80!white}{\cellcolor{gray!10!white} \textbf{\textit{Models w.o. RAG}}}} \\ 
\midrule
\multicolumn{1}{l|}{Qwen3-32B text-only} & 0.242 & 0.488 & 0.271 & 0.193 & 0.293 & 0.515 \\ 
\multicolumn{1}{l|}{Qwen2.5-VL-7B w.o. RAG} & 0.053 & 0.557 & 0.390 & 0.127 & 0.168 & 0.705 \\

\multicolumn{1}{l|}{GPT-4o w.o. RAG} & 0.109 & 0.748 & 0.144 & 0.197 & 0.332 & 0.472 \\ 

\multicolumn{1}{l|}{Qwen2.5-VL-32B w.o. RAG} & 0.200 & 0.621 & 0.179 & 0.159 & 0.319 & 0.521 \\ 

\multicolumn{1}{l|}{Llama 4 Maverick w.o. RAG} & \textbf{0.245} & 0.480 & 0.275 & \textbf{0.215} & 0.193 & 0.592\\

\midrule
\multicolumn{7}{c}{\textcolor{violet!80!white}{\cellcolor{gray!10!white} \textbf{\textit{Models Oracle}}}} \\ 
\midrule
\multicolumn{1}{l|}{Qwen2.5-VL-7B Oracle} & 0.406 & 0.490 & 0.104 & 0.456 & 0.241 & 0.303 \\

\multicolumn{1}{l|}{GPT-4o Oracle} & 0.678 & 0.141 & 0.181 & 0.538 & 0.271 & 0.191 \\ 

\multicolumn{1}{l|}{Llama 4 Maverick Oracle} & 0.601 & 0.350 & 0.049 & 0.524 & 0.192 & 0.284\\

\multicolumn{1}{l|}{Qwen2.5-VL-32B Oracle} & \textbf{0.874} & 0.061 & 0.066 & \textbf{0.643} & 0.312 & 0.045 \\ 
\bottomrule[1.5pt]
\end{tabular}
}
\end{wraptable}
\header{\method is high-quality and low contaminated that MLLMs still needs retrieval details to answer question correctly.}
As shown in Table \ref{tab:LMLLMs_Results}, \emph{state-of-the-art} MLLMs like \texttt{GPT-4o}, \texttt{Gemini}, and \texttt{Qwen} are able to make general responses without context, with 50\% to 70\% of responses being partially correct. Providing evidence pages to MLLMs substantially boosts accuracy, with 3x to 5x responses being completely correct compared to \emph{w.o.} RAG setting. This indicates that our benchmark is well-suited for evaluating the retrieval and synthesis components of RAG systems, as it clearly distinguishes context-grounded reasoning from a model's inherent knowledge. Notably, the robust performance of \texttt{Qwen2.5-VL} observed in the upper bound setting, which closely mirrors our benchmark curation pipeline, further suggesting the robustness and effectiveness of our pipeline in identifying correct evidence pages of queries.

\header{Document RAG frameworks bottleneck still lies on retrieval accuracy, where designing advanced strategies may help.}
Most frameworks strive to design complex information mining pipelines to extract maximum value from retrieved pages, yet tend to pay little attention to the retrieval stage itself. However, our experiments demonstrate strong correlations between retrieval accuracy and answer accuracy, as shown in Table \ref{tab:RAG_Framework_Results}. Equipped with a single MLLM pass, \texttt{Colqwen-gen} even partially outperforms \texttt{MDocAgent} on multi-hop queries, despite the latter seamlessly integrating multiple agents to provide final answers. This underscores the critical importance of optimizing the retrieval stage, potentially through finer-grained document preprocessing, exploiting the hierarchical and semantic structure of documents and developing more powerful or integrated embedding models.

\begin{table*}[!t]
\centering
\setlength{\tabcolsep}{4pt}
\caption{Average hit@5 performance across languages. Textual embedding model \texttt{Qwen3-Embedding-4B} even outperform multimodal models and frameworks in low-resource languages such as Arabic and Japanese, unveiling the challenge of current multimodal embedding models' generalizability to low-resource domains.}
\label{tab:summary_average_hop}
\resizebox{0.95\textwidth}{!}{
\begin{tabular}{l|c|c|c|c|c|c|c}
\toprule[1.5pt]
\textbf{Model \& Framework} & \textbf{Arabic} & \textbf{Chinese} & \textbf{English} & \textbf{French} & \textbf{Japanese} & \textbf{Spanish} & \textbf{Average} \\
\midrule
\multicolumn{8}{c}{\textcolor{violet!80!white}{\cellcolor{gray!10!white} \textbf{\textit{Text Embedding Models}}}}\\ 
\midrule
Qwen3-Embedding-4B & \textbf{0.685} & \underline{0.696} & 0.809 & 0.646 & \textbf{0.801} & 0.754 & \underline{0.732} \\
NV-Embed-v2 & 0.546 & 0.654 & \underline{0.819} & 0.660 & 0.662 & 0.698 & 0.673 \\
gte-Qwen2-7B-instruct & \underline{0.600} & 0.623 & 0.721 & 0.625 & 0.722 & 0.657 & 0.658 \\
bge-m3 & 0.331 & 0.451 & 0.727 & 0.453 & 0.486 & 0.489 & 0.490 \\
\midrule
\multicolumn{8}{c}{\textcolor{violet!80!white}{\cellcolor{gray!10!white} \textbf{\textit{Visual \& Multimodal Embedding Models}}}}\\ 
\midrule
colqwen2.5-3b-multilingual & 0.427 & 0.694 & \textbf{0.860} & \underline{0.702} & \underline{0.786} & \textbf{0.798} & 0.711 \\
vdr-2b-multi & 0.364 & 0.680 & 0.782 & 0.607 & 0.740 & 0.711 & 0.647 \\
jina-embeddings-v4 & 0.369 & 0.587 & 0.792 & 0.602 & 0.743 & 0.685 & 0.630 \\
gme-Qwen2-VL-7B-Instruct & 0.352 & 0.631 & 0.750 & 0.585 & 0.686 & 0.693 & 0.616 \\
colpali-v1.3 & 0.271 & 0.300 & 0.781 & 0.607 & 0.335 & 0.694 & 0.498 \\ \midrule
\multicolumn{8}{c}{\textcolor{violet!80!white}{\cellcolor{gray!10!white} \textbf{\textit{Document RAG System}}}}\\ 
\midrule
MDocAgent & 0.457 & 0.679 & 0.785 & 0.658 & 0.707 & 0.661 & 0.658 \\
ViDoRAG & 0.455 & 0.676 & 0.778 & 0.654 & 0.702 & 0.668 & 0.655 \\
M3DOCRAG & 0.377 & 0.579 & 0.704 & 0.558 & 0.621 & 0.599 & 0.573 \\
Colqwen-gen & 0.432 & \textbf{0.821} & 0.793 & \textbf{0.829} & 0.781 & \underline{0.771} & \textbf{0.738} \\
\bottomrule[1.5pt]
\end{tabular}
}
\end{table*}

\subsection{In-Depth Analysis}


\header{The overconfidence dilemma: trading trustworthiness for answers.}
To investigate the bottleneck in existing RAG frameworks, we breakdown each reponse of \texttt{M3DocRAG} and \texttt{MDocAgent} to analyze whether the error comes from retrieval or answering, and look into the trade-off between answering accuracy and the ability to identify insufficient information (also known as honesty \citep{gao2024honestllm}). 

\begin{wrapfigure}{r}{0.42\linewidth}
    \centering
    \vspace{-1em}
    \includegraphics[width=1\linewidth]{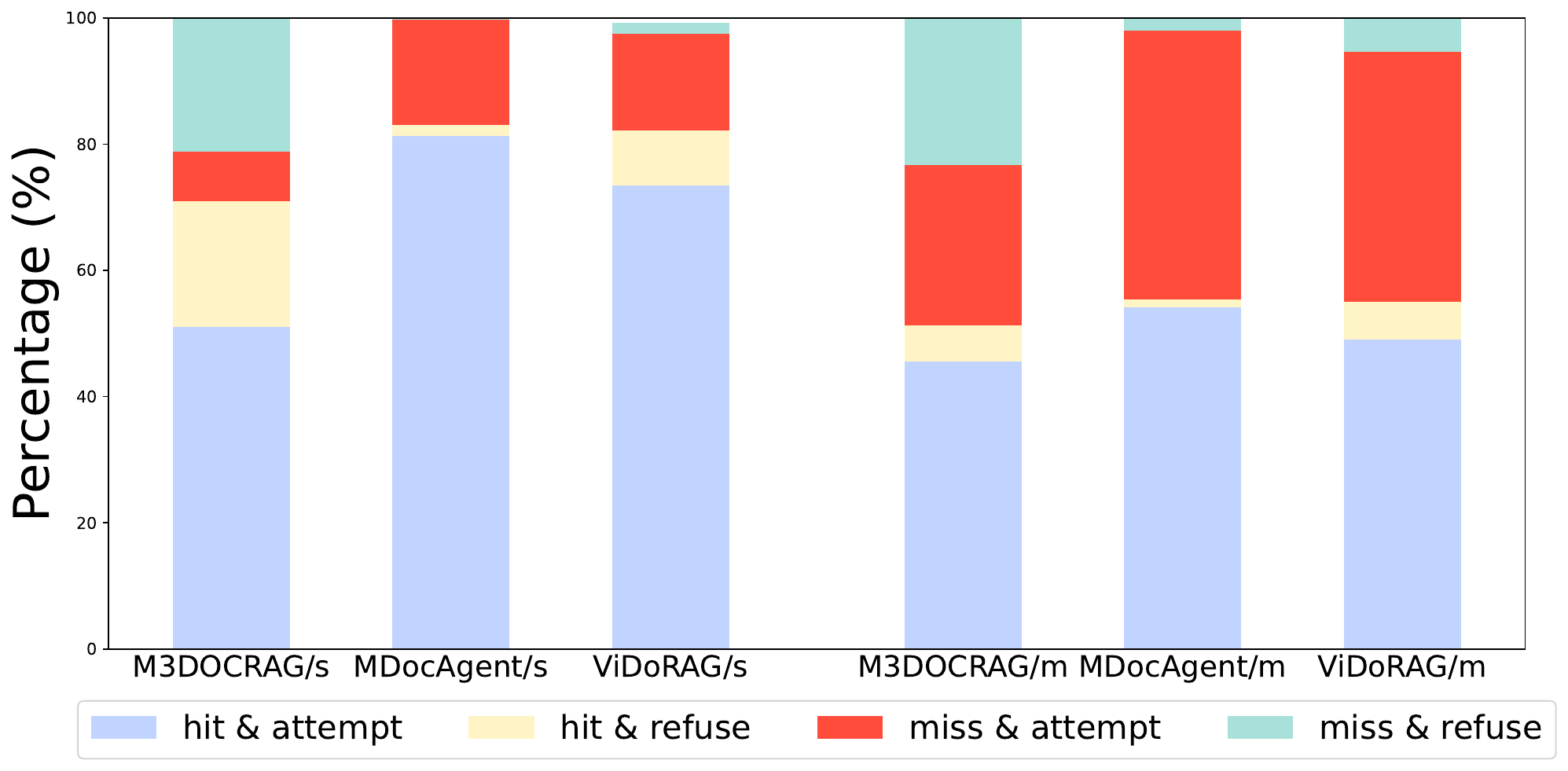}
    \vspace{-1em}
    \caption{Breakdown of retrieval and response model performance of frameworks under single (s) and multi-hop (m) queries. Our analysis reveals that performance drops on multi-hop questions are mainly due to retrieval failures that cause models to abstain from answering.}
    \label{fig:overconfidence analysis}
    \vspace{-1em}
\end{wrapfigure}
Figure \ref{fig:overconfidence analysis} reveals a striking divergence in agent behavior. Simpler agents like \texttt{M3DocRAG} adopt a cautious strategy, answering a lower proportion of queries with successfully retrieved context but reliably identifying retrieval failures and refusing to respond. In contrast, more complex agents like \texttt{MDocAgent} and \texttt{ViDoRAG} exhibit significant overconfidence. While they achieve higher accuracy on retrieval hits, they indiscriminately attempt to answer nearly every query, regardless of whether sufficient information was retrieved.
This frequently leads to speculative or entirely hallucinated content when evidence pages are missed.

This observation indicates that recent document RAG development has over-emphasized maximizing answer generation at the expense of \textit{``epistemic humility''}, \emph{i.e.}, the crucial skill of knowing what it doesn't know and admitting when an answer cannot be found. 
Consequently, we argue that future research should pursue more trustworthy RAG frameworks where identifying informational gaps is as valued as accuracy \citep{zhou2025trustrag, huang2025trustworthiness}. 

\header{Inference patterns of MLLMs as response model.}
We also observe different answering strategy in MLLMs in Table \ref{tab:RAG_Framework_Results} and Appendix. When directly provided with a multi-hop query, response model tend not to process them hop-by-hop.  On the contrary, they first collect signature information---the most distinguishing or identifiable pieces---from the various hops. Following this, models tend to perform a direct inclusion based elimination to arrive the final answer. This mechanism differentiates significantly from our expectation of how models might sequentially solve multi-hop queries. This provides a compelling point of view: merely increasing the number of hops may not increase its difficulty.
A case study in Appendix \ref{inference case study} reveals that MLLMs do not process multi-hop queries step-by-step as expected. Instead, they gather key signature information from each hop and use inclusion-based elimination to find the answer. This challenges the assumption that more hops always increase difficulty, suggesting further investigation is needed.

\begin{wrapfigure}{r}{0.4\linewidth}
    \centering
    \vspace{-1em}
    \includegraphics[width=1\linewidth]{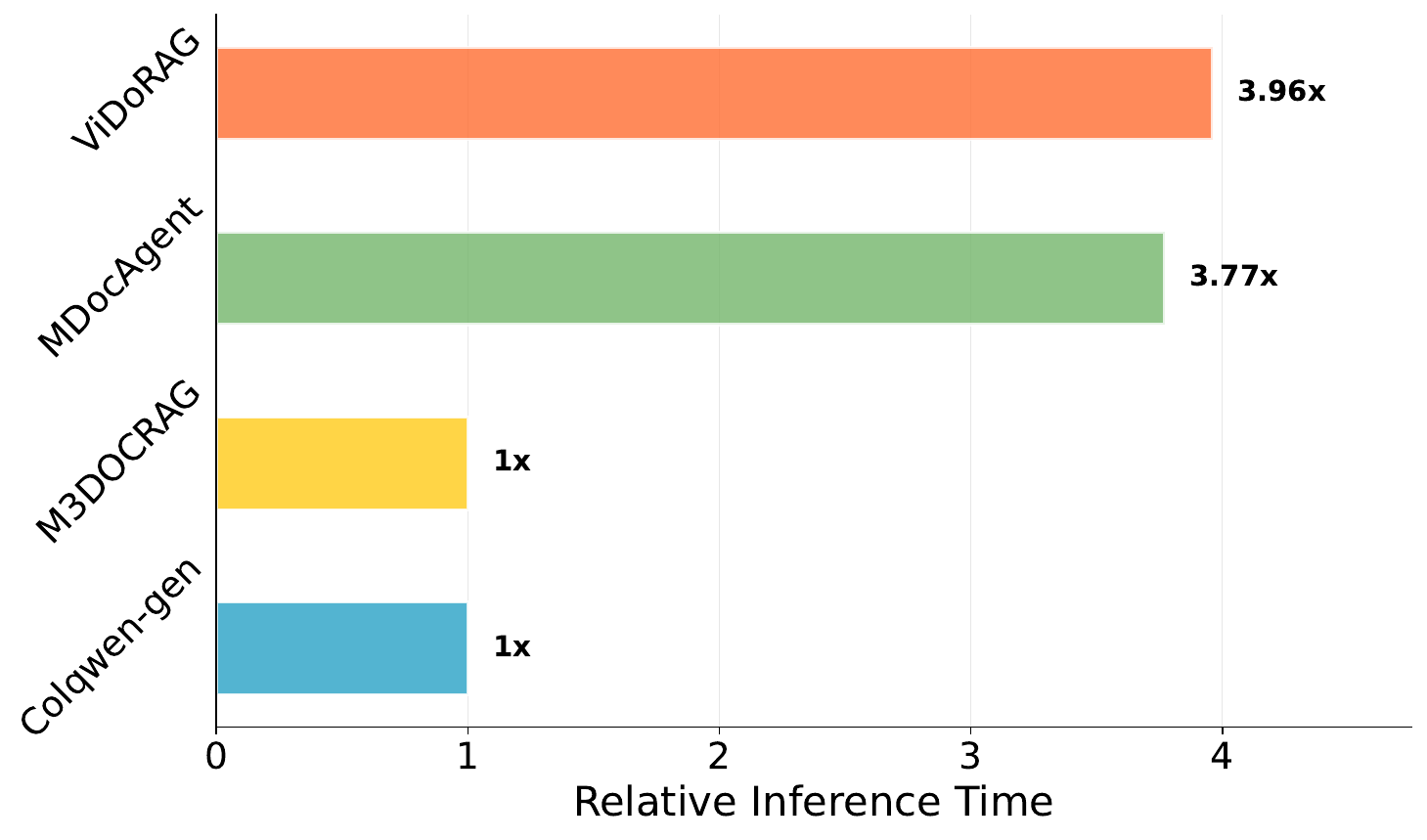}
    \vspace{-1em}
    \caption{Time efficiency of each document RAG system.}
    \label{fig:latency ablation}
    \vspace{-2em}
\end{wrapfigure}

\header{Time efficiency of frameworks.}
Agent efficiency is as important a metric as effectiveness. Since API completion times may vary across models, Figure \ref{fig:latency ablation} reports the normalized time efficiency of the evaluated frameworks. Both \texttt{MDocAgent} and \texttt{ViDoRAG} employ a sequential agent coordination pattern, which significantly increases their inference time. Note that \texttt{ViDoRAG} dynamically controls the generation process, so we report the lower and upper bound theoretical time efficiency of \texttt{ViDoRAG} estimated by API call times.

\begin{table*}[!t]
\centering
\setlength{\tabcolsep}{4pt}
\resizebox{0.72\textwidth}{!}{
\begin{tabular}{l|ccc|ccc|ccc}
\toprule[1.5pt]
\multirow{2}{*}{\textbf{Framework}} & \multicolumn{3}{c|}{\textbf{Figure}} & \multicolumn{3}{c|}{\textbf{Text}} & \multicolumn{3}{c}{\textbf{Table}} \\
\cmidrule{2-10}
 & \yes & \half & \multicolumn{1}{c|}{\no}
& \yes & \half & \multicolumn{1}{c|}{\no} & \yes & \half & \multicolumn{1}{c|}{\no} \\
\midrule
\multicolumn{1}{c|}{MDocAgent} & \textbf{0.692} & 0.106 & \multicolumn{1}{c|}{0.202} & \textbf{0.665} & 0.114 & \multicolumn{1}{c|}{0.221} & \textbf{0.547} & 0.148 & 0.305 \\
\multicolumn{1}{c|}{ViDoRAG} & 0.609 & 0.122 & \multicolumn{1}{c|}{0.269} & 0.557 & 0.138 & \multicolumn{1}{c|}{0.305} & 0.489 & 0.151 & 0.360 \\
\multicolumn{1}{c|}{M3DOCRAG} & 0.456 & 0.127 & 0.418 & 0.512 & 0.110 & \multicolumn{1}{c|}{0.378} & 0.371 & 0.114 & 0.515 \\
\multicolumn{1}{c|}{Colqwen-gen} & 0.657 & 0.186 & 0.157 & 0.619 & 0.114 & \multicolumn{1}{c|}{0.268} & 0.545 & 0.104 & 0.351 \\
\bottomrule[1.5pt]
\end{tabular}
}
\caption{Performance of multimodal document RAG system across different modalities.}
\label{tab:framework_answer_modality_qa}
\end{table*}
\section{Related Work}
\paragraph{Multimodal Document Retrieval.}
Different from traditional text retrieval \citep{zhao2024dense,blandon2025memerag}, documents \citep{masry2022chartqa,tanaka2023slidevqa} often contain multimodal information, which may be time consuming and would cause information loss when directly parsed to text. Therefore, recent works have dedicated great effort to improve the accuracy and efficiency of document retrieval with MLLMs. One line of work adopts high quality synthetic data \citep{zhang2024gme, chen2025mme5}, hardness aware training \citep{lan2025llave, lee2024nv} and retrieval-optimized network architectures \citep{faysse2024colpali} for more precise embedding models. Another line of work leverages LLM/MLLM agentic flows to process different modalities in parallel \citep{han2025mdocagent,ouyang2025nvagent} and perform iterative inference steps for more grounded and informative answers  \citep{wang2025vidorag}. 

\paragraph{Document RAG Benchmarks.}
The increasing attention on Document RAG \citep{ma2024mmlongbench} and VQA \citep{mathew2021docvqa} necessitates comprehensive multimodal retrieval benchmarks. Common practices often use VQA dataset queries \citep{friel2024ragbench, faysse2024colpali, cho2024m3docrag,fu2025livevqa}, but these are document-specific and lack information for global retrieval. Other benchmarks \citep{wang2025vidorag, dong2025mmdocir} craft informative queries from single pages, yet often only mark that single page as relevant, ignoring other potential matches and risking evaluation inaccuracies.
Some recent benchmarks \citep{mace2025vidore, wasserman2025real,xu2025mmrag} have identified contextual gaps between artificial and realistic queries, and strive to provide evaluations that fully reflect real use scenarios. 

\section{Conclusion}
We introduce \method, a large-scale, multimodal, multilingual benchmark designed to reflect realistic retrieval-augmented generation scenarios, overcoming limitations of prior work with validated chain-of-evidence and comprehensive assessment. Evaluations of leading embedding models and RAG frameworks reveal several crucial bottlenecks. We hope our fully open-sourced code, framework and dataset establish a strong foundation for document RAG system.

\section*{Limitations}
We leverage LLMs to revise our paper and serving as metrics in our evaluation. We include human-annotation in Appendix \ref{appendix: human} to validate the LLM-as-a-Judge process.

Despite \method's comprehensive design, several limitations constrain its scope and applicability. First, although our benchmark provides the broadest coverage across 6 languages and 4 document types compared to previous document RAG benchmarks, it potentially misses important linguistic communities and specialized domains such as legal or medical documents that present unique RAG challenges. Second, our query synthesis pipeline relies heavily on large language models (primarily \texttt{GPT-4o} and \texttt{Qwen2.5-VL-32B-Instruct}), thereby introducing systematic biases in question generation patterns that may not reflect the full diversity of human information-seeking behavior.

\bibliography{custom}

\newpage
\appendix

\section{Detailed Benchmark Construction}
\begin{wrapfigure}{r}{0.4\linewidth}
    \centering
    \includegraphics[width=\linewidth]{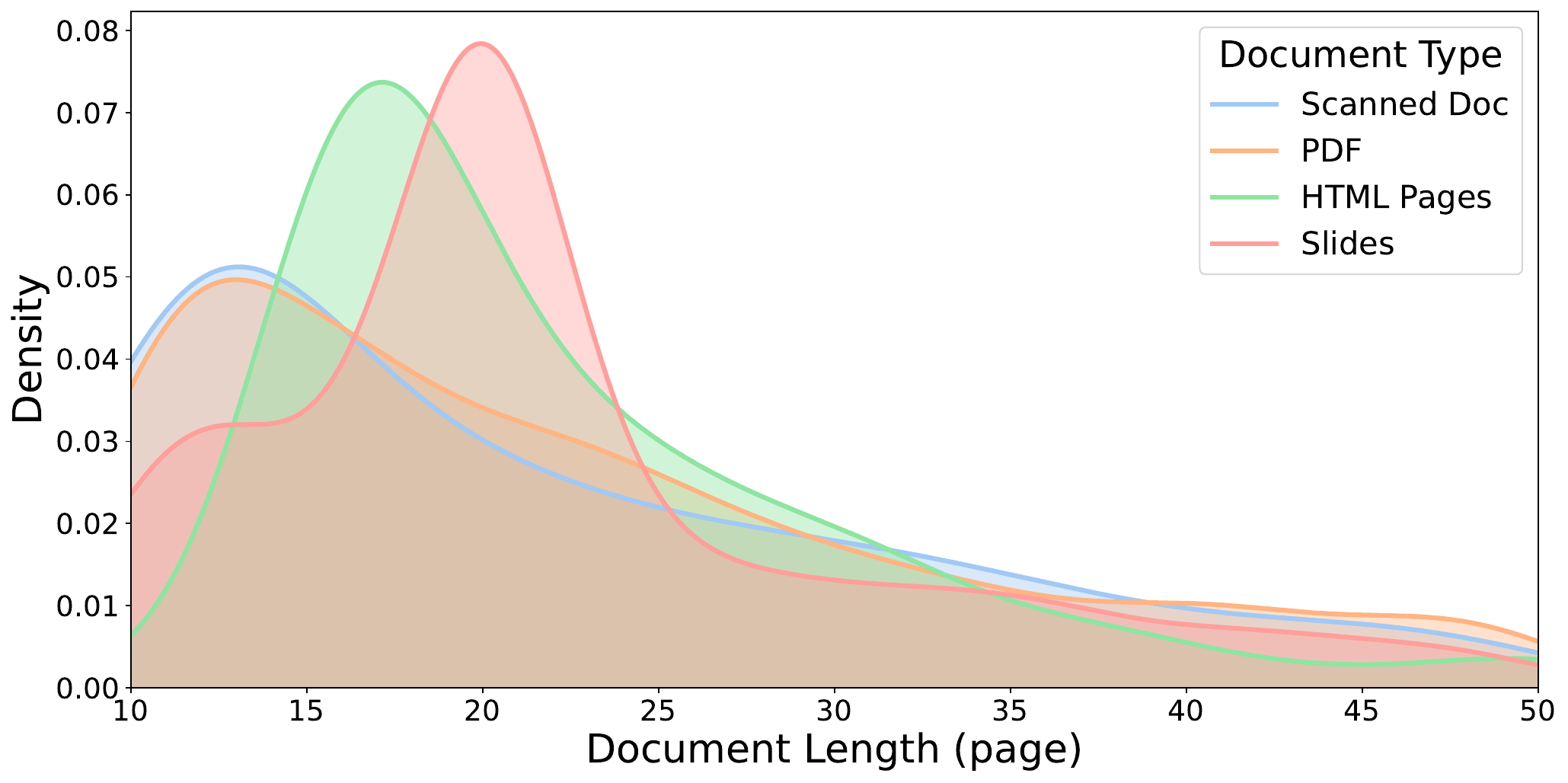}
    \caption{KDE distribution of document lengths by document type.}
    \label{fig:doc len}
    \vspace{-1em}
\end{wrapfigure}

\header{Step 1: Data Collection and Preprocessing.}
We collected rich documents from various sources, including  PDFs, scanned documents, slides and HTML pages  in multiple languages. We then selected documents with a length between 10-50 pages and eliminated irrelevant content. Using tools such as Docling and MinerU, we split each document into text, tables, and figures, ensuring that each modality was processed separately. We further used GPT-4o to filter out low-quality or irrelevant data chunks and retain only important content for query generation. Finally, the detailed data for our benchmark is shown in Table~\ref{dataset statistics}, Table~\ref{tab:dataset summary} and Figure~\ref{fig:doc len}.

\header{Step 2: One-Hop Query Synthesis.}
To generate single-hop queries, we began by utilizing parsed document chunks. For textual content, we focused on identifying key entities, data, and concepts to craft self-contained questions. Meanwhile, for visual elements, we prioritized understanding critical components and patterns. Furthermore, to handle ambiguous queries that yielded more than five ground truth pages, we designated a specific evidence page and prompted the model to analyze it in contrast with the other pages, thereby adding a distinguishing detail to refine the query and ensure its specificity. Finally, each query was meticulously reviewed to ensure clarity, relevance, and direct answerability from the document.

\begin{wraptable}{r}{0.4\linewidth}
\centering
\vspace{-1em}
\caption{\method statistics.}
\label{dataset statistics}

\setlength{\tabcolsep}{6pt}
\begin{tabular}{lc}
\toprule[1.5pt]
\textbf{Statistic} & \textbf{Number} \\
\midrule
\textbf{Documents} & 3276 \\
\quad Languages & 6 \\
\quad Avg. pages per doc & 22.3 \\
\quad Avg. words per page & 289.9 \\
\quad Avg. tables per page & 0.397 \\
\quad Avg. figures per page & 1.078 \\
\midrule
\textbf{Total Questions} & 5168 \\
\quad Single-hop questions & 2500 \\
\quad \quad Avg. evidence pages & 2.91 \\
\quad Multi-hop questions & 2668 \\
\quad \quad Avg. hops & 2.78 \\
\bottomrule
\end{tabular}

\vspace{-1em}
\end{wraptable}
\header{Step 3: Multi-Hop Query Synthesis.}
To address the challenge of multi-hop questions that require reasoning across multiple steps, we employed a knowledge graph-based approach. Using LightRAG, we constructed knowledge graphs for each document, extracting entities and relations from both textual and visual content. These graphs leveraged the connectivity of nodes and edges to represent the relationships between different pieces of information. We selected query paths through these graphs, starting with high-dimensional entities and expanding paths based on content depth and connectivity. Subqueries were synthesized for each hop and iteratively combined into a coherent multi-hop query. Each query was rigorously checked to ensure its logical coherence, unique answers, and overall relevance to the document.

\header{Step 4: Post-Processing.}
After query synthesis, the generated queries underwent model filtering using advanced models to ensure they met quality standards, removing poorly structured or irrelevant queries. Evidence pages for both single-hop and multi-hop queries were annotated by MLLMs like Qwen2.5-32B-VL, ensuring that supporting evidence was accurately identified. Finally, human annotators reviewed the filtered queries and evidence pages for accuracy and consistency. Human refinement addressed potential discrepancies, thereby enhancing the overall reliability of the benchmark.. This comprehensive post-processing approach ensured the robustness of the benchmark and its suitability for evaluating multimodal document retrieval systems.\\


\section{Human Annotation Details}
\label{appendix: human}
\begin{figure}[thb]
    \centering
    \includegraphics[width=1\linewidth]{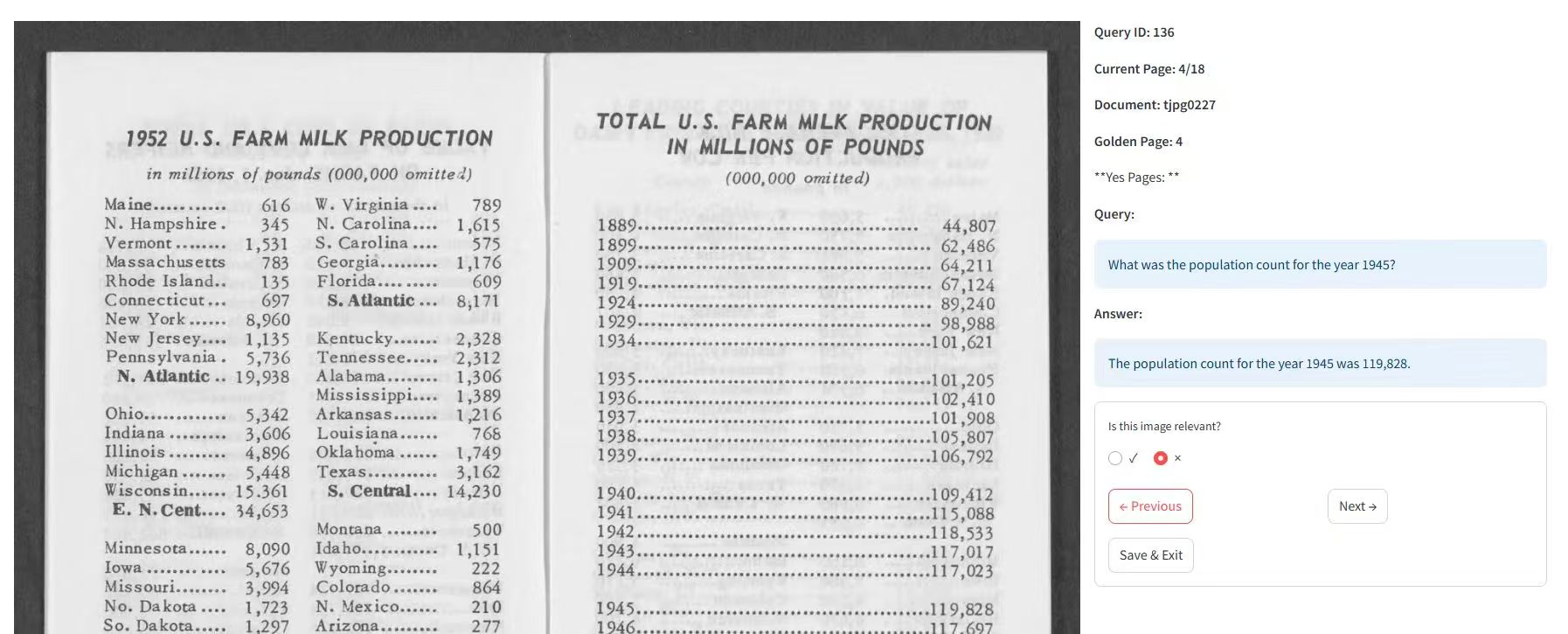}
    \caption{Human Annotation UI.}
    \label{fig:human}
\end{figure}

The annotation is conducted by 5 authors of this paper and 1 volunteers independently. As acknowledged, the diversity of annotators plays a crucial role in reducing bias and enhancing the reliability of the benchmark. These annotators have knowledge in this domain, with different genders, ages, and educational backgrounds. To ensure the annotators can proficiently mark the data, we provide them with detailed tutorials, teaching them how to evaluate model responses more objectively. The annotation UI is shown in Figure~\ref{fig:human}.

To guarantee the quality and reliability of our benchmark, we implemented a comprehensive human verification process for both the generated queries and the ground truth evidence labels. First, every query underwent a rigorous human review after an initial automated screening. This step focused on ensuring the questions were high-quality, realistic, and compliant with our generation standards, culminating in a 97\% agreement rate between the model's judgments and human annotators on a 200-item subsample. In parallel, the evidence page labels were subjected to a thorough human audit, which revealed an initial 92\% agreement rate between the automated labels and human judgment. Our annotators then meticulously resolved the remaining discrepancies. This dual-layered human verification process ensures that both the queries and their ground truth labels are of high fidelity, establishing a robust foundation for evaluation.

\section{Detailed Experiment Settings}
\subsection{Experimental Methodology}

\subsubsection{Experiment 1: Evaluation of Embedding Models}

To assess the recall capabilities of state-of-the-art embedding models, we conducted experiments on both multimodal and text embedding models. For multimodal embedding models, we directly embedded the visual and textual content of the documents. In contrast, for text embedding models, we first applied OCR to extract text from images and then embedded the extracted text. Additionally, for images and tables, we used a Vision-Language Model (VLM) to generate descriptive captions, which were subsequently embedded. Finally, we also counted the hit@5 of these embed models in single-hop, 2-hop, and 3-hop queries in different languages, as shown in Table~\ref{tab:summary_single_hop}, Table~\ref{tab:summary_2_hop} and Table~\ref{tab:summary_3_hop}.The experimental results highlight a strong competition between two leading models: the multimodal colqwen2.5-3b-multilingual and the text-based Qwen3-Embedding-4B. Overall, colqwen2.5-3b performs slightly better than Qwen3-Embedding-4B. However, in certain languages, such as Arabic, the text-based Qwen3-Embedding-4B shows superior performance.

\begin{table*}[!htb]
\centering
\setlength{\tabcolsep}{4pt}
\caption{Single-Hop hit@5 performance of embedding model across languages.}
\label{tab:summary_single_hop}
\resizebox{0.95\textwidth}{!}{
\begin{tabular}{l|c|c|c|c|c|c}
\toprule[1.5pt]
\textbf{Model} & \textbf{Arabic} & \textbf{Chinese} & \textbf{English} & \textbf{French} & \textbf{Japanese} & \textbf{Spanish} \\
\midrule
\multicolumn{7}{c}{\textcolor{violet!80!white}{\cellcolor{gray!10!white} \textbf{\textit{Text Embedding Models}}}}\\
\midrule
Qwen3-Embedding-4B & \textbf{0.668} & \textbf{0.791} & \textbf{0.963} & \textbf{0.883} & \textbf{0.872} & \textbf{0.881} \\
NV-Embed-v2 & 0.495 & 0.636 & 0.933 & 0.820 & 0.653 & 0.814 \\
gte-Qwen2-7B-instruct & 0.639 & 0.667 & 0.881 & 0.808 & 0.766 & 0.814 \\
bge-m3 & 0.361 & 0.535 & 0.919 & 0.614 & 0.508 & 0.631 \\
\midrule 
\multicolumn{7}{c}{\textcolor{violet!80!white}{\cellcolor{gray!10!white} \textbf{\textit{Visual \& Multimodal Embedding Models}}}}\\
\midrule
colqwen2.5-3b-multilingual & \textbf{0.466} & \textbf{0.841} & \textbf{0.983} & \textbf{0.925} & \textbf{0.872} & \textbf{0.927} \\
vdr-2b-multi & 0.413 & 0.771 & 0.964 & 0.838 & 0.781 & 0.895 \\
jina-embeddings-v4 & 0.394 & 0.760 & 0.948 & 0.835 & 0.827 & 0.895 \\
gme-Qwen2-VL-7B-Instruct & 0.418 & 0.721 & 0.942 & 0.832 & 0.766 & 0.836 \\
colpali-v1.3 & 0.255 & 0.333 & 0.962 & 0.799 & 0.362 & 0.825 \\
\bottomrule[1.5pt]
\end{tabular}
}
\end{table*}
\begin{table*}[!htb]
\centering
\setlength{\tabcolsep}{4pt}
\caption{2-Hop hit@5 performance of embedding model across languages.}
\label{tab:summary_2_hop}
\resizebox{0.95\textwidth}{!}{
\begin{tabular}{l|c|c|c|c|c|c}
\toprule[1.5pt]
\textbf{Model} & \textbf{Arabic} & \textbf{Chinese} & \textbf{English} & \textbf{French} & \textbf{Japanese} & \textbf{Spanish} \\
\midrule
\multicolumn{7}{c}{\textcolor{violet!80!white}{\cellcolor{gray!10!white} \textbf{\textit{Text Embedding Models}}}}\\
\midrule
Qwen3-Embedding-4B & \textbf{0.730} & 0.731 & 0.737 & 0.435 & \textbf{0.782} & \textbf{0.754} \\
NV-Embed-v2 & 0.635 & \textbf{0.827} & \textbf{0.796} & \textbf{0.571} & 0.709 & 0.678 \\
gte-Qwen2-7B-instruct & 0.587 & 0.673 & 0.664 & 0.506 & 0.691 & 0.593 \\
bge-m3 & 0.222 & 0.462 & 0.620 & 0.325 & 0.436 & 0.432 \\
\midrule 
\multicolumn{7}{c}{\textcolor{violet!80!white}{\cellcolor{gray!10!white} \textbf{\textit{Visual \& Multimodal Embedding Models}}}}\\
\midrule
colqwen2.5-3b-multilingual & \textbf{0.556} & 0.673 & \textbf{0.796} & \textbf{0.539} & \textbf{0.818} & \textbf{0.797} \\
vdr-2b-multi & 0.365 & \textbf{0.692} & 0.701 & 0.435 & 0.745 & 0.661 \\
jina-embeddings-v4 & 0.413 & 0.538 & 0.737 & 0.416 & 0.764 & 0.585 \\
gme-Qwen2-VL-7B-Instruct & 0.349 & 0.673 & 0.650 & 0.429 & 0.764 & 0.686 \\
colpali-v1.3 & 0.381 & 0.423 & 0.664 & 0.461 & 0.309 & 0.669 \\
\bottomrule[1.5pt]
\end{tabular}
}
\end{table*}
\begin{table*}[!htb]
\centering
\setlength{\tabcolsep}{4pt}
\caption{3-Hop hit@5 performance embedding model across languages.}
\label{tab:summary_3_hop}
\resizebox{0.95\textwidth}{!}{
\begin{tabular}{l|c|c|c|c|c|c}
\toprule[1.5pt]
\textbf{Model} & \textbf{Arabic} & \textbf{Chinese} & \textbf{English} & \textbf{French} & \textbf{Japanese} & \textbf{Spanish} \\
\midrule
\multicolumn{7}{c}{\textcolor{violet!80!white}{\cellcolor{gray!10!white} \textbf{\textit{Text Embedding Models}}}}\\
\midrule
Qwen3-Embedding-4B & \textbf{0.658} & \textbf{0.567} & \textbf{0.728} & \textbf{0.620} & \textbf{0.750} & \textbf{0.628} \\
NV-Embed-v2 & 0.507 & 0.500 & \textbf{0.728} & 0.588 & 0.625 & 0.603 \\
gte-Qwen2-7B-instruct & 0.575 & 0.529 & 0.617 & 0.560 & 0.708 & 0.564 \\
bge-m3 & 0.411 & 0.356 & 0.643 & 0.419 & 0.514 & 0.404 \\
\midrule 
\multicolumn{7}{c}{\textcolor{violet!80!white}{\cellcolor{gray!10!white} \textbf{\textit{Visual \& Multimodal Embedding Models}}}}\\
\midrule
colqwen2.5-3b-multilingual & 0.260 & 0.567 & \textbf{0.802} & \textbf{0.641} & 0.667 & \textbf{0.669} \\
vdr-2b-multi & \textbf{0.315} & \textbf{0.577} & 0.681 & 0.549 & \textbf{0.694} & 0.576 \\
jina-embeddings-v4 & 0.301 & 0.462 & 0.691 & 0.556 & 0.639 & 0.576 \\
gme-Qwen2-VL-7B-Instruct & 0.288 & 0.500 & 0.657 & 0.495 & 0.528 & 0.557 \\
colpali-v1.3 & 0.178 & 0.144 & 0.717 & 0.562 & 0.333 & 0.587 \\
\bottomrule[1.5pt]
\end{tabular}
}
\end{table*}

\begin{table*}[!htb]
\centering
\setlength{\tabcolsep}{4pt}
\caption{Single-hop hit@5 performance of multimodal document RAG frameworks across languages.}
\label{tab:framwork_single_hit}
\resizebox{0.75\textwidth}{!}{
\begin{tabular}{l|c|c|c|c|c|c}
\toprule[1.5pt]
\textbf{Framework} & \textbf{Arabic} & \textbf{Chinese} & \textbf{English} & \textbf{French} & \textbf{Japanese} & \textbf{Spanish} \\
\midrule
MDocAgent & 0.413 & 0.786 & 0.942 & 0.808 & 0.775 & 0.841 \\
ViDoRAG & 0.420 & 0.769 & 0.945 & 0.791 & 0.773 & 0.847 \\
M3DOCRAG & \textbf{0.475} & 0.621 & 0.856 & 0.627 & 0.658 & 0.690 \\
Colqwen-gen & 0.466 & \textbf{0.841} & \textbf{0.983} & \textbf{0.925} & \textbf{0.872} & \textbf{0.927} \\
\bottomrule[1.5pt]
\end{tabular}
}

\end{table*}

\begin{table*}[!htb]
\centering
\setlength{\tabcolsep}{4pt}
\caption{Multi-hop hit@5 performance of document RAG frameworks across languages.}
\label{tab:framework_multi_hit}
\resizebox{0.75\textwidth}{!}{
\begin{tabular}{l|c|c|c|c|c|c}
\toprule[1.5pt]
\textbf{Framework} & \textbf{Arabic} & \textbf{Chinese} & \textbf{English} & \textbf{French} & \textbf{Japanese} & \textbf{Spanish} \\
\midrule
MDocAgent & \textbf{0.500} & 0.571 & \textbf{0.628} & 0.507 & 0.638 & 0.480 \\
ViDoRAG & 0.489 & \textbf{0.582} & 0.611 & 0.517 & 0.631 & 0.488 \\
M3DOCRAG & 0.279 & 0.537 & 0.551 & 0.489 & 0.583 & 0.508 \\
Colqwen-gen & 0.397 & 0.801 & 0.603 & \textbf{0.732} & \textbf{0.690} & \textbf{0.614} \\
\bottomrule[1.5pt]
\end{tabular}
}
\end{table*}

\begin{table*}[!htb]
\centering
\setlength{\tabcolsep}{4pt}
\caption{Single-hop hit@5 performance of document RAG framework across document types.}
\label{tab:framework_single_doc_type}
\resizebox{0.6\textwidth}{!}{
\begin{tabular}{l|c|c|c|c}
\toprule[1.5pt]
\textbf{Framework} & \textbf{PDF} & \textbf{Slides} & \textbf{HTML Pages} & \textbf{Scanned Doc} \\
\midrule
MDocAgent & 0.781 & 0.858 & 0.842 & 0.636 \\
ViDoRAG & 0.776 & 0.852 & 0.834 & 0.638 \\
M3DOCRAG & 0.778 & 0.592 & 0.681 & 0.566 \\
Colqwen-gen & \textbf{0.905} & \textbf{0.873} & \textbf{0.901} & \textbf{0.808}\\
\bottomrule[1.5pt]
\end{tabular}
}

\end{table*}

\begin{table*}[!htb]
\centering
\setlength{\tabcolsep}{4pt}
\caption{Multi-hop hit@5 performance of document RAG frameworks across document types.}
\label{tab:framework_multi_doc_type}
\resizebox{0.6\textwidth}{!}{
\begin{tabular}{l|c|c|c|c}
\toprule[1.5pt]
\textbf{Framework} & \textbf{PDF} & \textbf{Slides} & \textbf{HTML Pages} & \textbf{Scanned Doc} \\
\midrule
MDocAgent & 0.567 & 0.570 & 0.538 & 0.536 \\
ViDoRAG & 0.572 & 0.552 & 0.532 & 0.540 \\
M3DOCRAG & 0.560 & 0.538 & 0.505 & 0.321 \\
Colqwen-gen & \textbf{0.711} & \textbf{0.689} & \textbf{0.669} & \textbf{0.726}\\
\bottomrule[1.5pt]
\end{tabular}
}
\end{table*}

\begin{table*}[!htb]
\centering
\setlength{\tabcolsep}{4pt}
\caption{Single-hop hit@5 performance of document RAG system across modalities.}
\label{tab:framework_single_modality}
\resizebox{0.4\textwidth}{!}{
\begin{tabular}{l|c|c|c}
\toprule[1.5pt]
\textbf{Framework} & \textbf{Figure} & \textbf{Text} & \textbf{Table} \\
\midrule
MDocAgent & 0.917 & 0.820 & 0.731 \\
ViDoRAG & \textbf{0.920} & 0.802 & 0.738 \\
M3DOCRAG & 0.629 & 0.849 & 0.641 \\
Colqwen-gen & 0.877 & \textbf{0.857} & \textbf{0.917}\\
\bottomrule[1.5pt]
\end{tabular}
}
\end{table*}

\begin{table*}[!htb]
\centering
\setlength{\tabcolsep}{4pt}
\caption{Multi-hop hit@5 performance of document RAG frameworks across modalities.}
\label{tab:framework_Multi_modality}
\resizebox{0.4\textwidth}{!}{
\begin{tabular}{l|c|c|c}
\toprule[1.5pt]
\textbf{Framework} & \textbf{Figure} & \textbf{Text} & \textbf{Table} \\
\midrule
MDocAgent & 0.615 & 0.557 & 0.559 \\
ViDoRAG & 0.606 & 0.561 & 0.552 \\
M3DOCRAG & 0.564 & 0.522 & 0.509 \\
Colqwen-gen & 0.704 & 0.706 &  0.688 \\
\bottomrule[1.5pt]
\end{tabular}
}

\end{table*}

\begin{table*}[!htb]
\centering
\setlength{\tabcolsep}{4pt}
\caption{Performance of multimodal document RAG system across different document types.}
\label{tab:framework_multi_doc_type_qa}
\resizebox{0.95\textwidth}{!}{
\begin{tabular}{l|ccc|ccc|ccc|ccc}
\toprule[1.5pt]
\multirow{2}{*}{\textbf{Framework}} & \multicolumn{3}{c|}{\textbf{PDF}} & \multicolumn{3}{c|}{\textbf{Slides}} & \multicolumn{3}{c|}{\textbf{HTML Pages}} & \multicolumn{3}{c}{\textbf{Scanned Doc}} \\
\cmidrule{2-13}
 & \yes & \half & \multicolumn{1}{c|}{\no}
& \yes & \half & \multicolumn{1}{c|}{\no} & \yes & \half & \multicolumn{1}{c|}{\no} & \yes & \half & \multicolumn{1}{c|}{\no} \\
\midrule
MDocAgent & \textbf{0.661} & 0.123 & 0.216 & \textbf{0.675} & 0.118 & 0.206 & \textbf{0.631} & 0.149 & 0.220 & \textbf{0.625} & 0.130 & 0.245 \\
ViDoRAG & 0.534 & 0.131 & 0.335 & 0.541 & 0.127 & 0.332 & 0.522 & 0.152 & 0.325 & 0.539 & 0.147 & 0.314 \\
M3DOCRAG & 0.488 & 0.121 & 0.391 & 0.472 & 0.107 & 0.421 & 0.454 & 0.084 & 0.462 & 0.354 & 0.132 & 0.514 \\
Colqwen-gen & 0.636 & 0.130 & 0.235 & 0.619 & 0.062 & 0.32 & 0.575 & 0.156 & 0.269 & 0.560 & 0.180 & 0.260\\
\bottomrule[1.5pt]
\end{tabular}
}
\end{table*}

\subsubsection{Experiment 2: Performance of LLMs with and without Golden Pages}

We tested the performance of advanced Large Language Models (LLMs) and Vision-Language Models (VLMs), including Qwen3-32B, GPT-4o, Llama 4 Maverick, Qwen2.5-VL-7B and Qwen2.5-VL-32B, in two different contexts. First, we evaluated their performance when provided with the exact 5 golden pages as context, which allowed us to assess their ability to generate accurate answers with direct access to relevant information. Second, we tested their performance without the golden pages, relying solely on their inherent knowledge and reasoning capabilities. This dual evaluation provided insights into how these models perform in both ideal and more challenging conditions.

\subsubsection{Experiment 3: RAG Frameworks Evaluation}

To further evaluate the effectiveness of Retrieval-Augmented Generation (RAG) frameworks, we conducted experiments using MDocAgent, ViDoRAG, and M3DOCRAG. In addition to these systems, we established a baseline named Colqwen-Gen, which pairs the retrieval results from Colqwen-3B with the generation capabilities of GPT-4o. We tested both the recall capabilities of these frameworks in retrieving relevant documents and their ability to generate accurate answers based on the retrieved information. In addition, we also analyzed the Time efficiency of Frameworks,as shown in Figure \ref{fig:latency ablation}.We also counted and analyzed the hit@5 and answer performance of these frameworks in different languages, document types, and modalities, as shown in Table~\ref{tab:framwork_single_hit}, Table~\ref{tab:framework_multi_hit}, Table~\ref{tab:framework_single_doc_type}, 
Table~\ref{tab:framework_multi_doc_type},
Table~\ref{tab:framework_single_modality},
Table~\ref{tab:framework_Multi_modality}
Table~\ref{tab:framework_answer_modality_qa} and Table~\ref{tab:framework_multi_doc_type_qa}.

\section{Prompt Templates}

Full prompt templates and examples are provided in Figure~\ref{fig:pilot-study-prompt}, Figure~\ref{fig:visual-filter-system-prompt}, Figure~\ref{fig:singlehop-qa-prompt}, Figure~\ref{fig:singlehop-qa-image-prompt}, Figure~\ref{fig:singlehop-text-filter-prompt}, Figure~\ref{fig:singlehop-image-filter-prompt}, Figure~\ref{fig:table-to-text-caption-prompt}, Figure~\ref{fig:figure-to-text-caption-prompt}, Figure~\ref{fig:relationship-evaluation-selection-prompt}, Figure~\ref{fig:relationship-selection-prompt}, Figure~\ref{fig:step-qa-generation-prompts}, Figure~\ref{fig:multi-hop-question-chaining}, Figure~\ref{fig:multihop-qa-filter-prompt}, Figure~\ref{fig:page-gt-prompt}, Figure~\ref{fig:Question-Refinement-Prompt}, Figure~\ref{fig:answer generation prompt oracle} and Figure~\ref{fig:qa-judge-prompt}.

\newtcolorbox{promptbox}[1]{
    colback=white,            
    colframe=black,           
    fonttitle=\bfseries,      
    title=#1,                 
    boxsep=5pt,               
    arc=1.5mm, auto outer arc,                
    left=10pt, right=10pt, top=5pt, bottom=5pt, 
    boxrule=0.8pt,             
    width=\linewidth
}
\section{Case Study}
A detailed case study is shown in Figure~\ref{fig:case_study_MLLM} .

\label{inference case study}


\begin{figure*}[thb]
    \begin{promptbox}{Case Study:}

\textbf{Multi hop Question:}

What disease does the entity that launched its \hl{first nationwide public service campaign in 1980} implement programs to prevent, which causes \hl{blindness in children due to vitamin A deficiency?} \\
\textbf{MLLM response:}

The disease that the entity likely implemented programs to prevent, which \hl{causes blindness in children due to vitamin A deficiency, is Xerophthalmia.} \textbf{Note:} This discussion is missing the step of identifying the first nationwide public service campaign in 1980 that implemented such prevention programs.

While pinpointing a single entity that launched its \hl{first nationwide public service campaign} specifically in 1980 in the United States focused solely on preventing xerophthalmia through vitamin A sufficiency proves challenging based on available search information, the 1980s were a crucial period for understanding and combating vitamin A deficiency and its impact on childhood blindness and mortality.

Research in the 1980s, significantly contributed to by organizations like Helen Keller Intl in collaboration with researchers such as Dr. Alfred Sommer, highlighted the profound impact of vitamin A supplementation in preventing not only blindness but also reducing childhood deaths from common illnesses. This groundbreaking evidence galvanized international health organizations and NGOs to expand their efforts in vitamin A distribution and public health initiatives in areas where deficiency was prevalent.

Therefore, while a specific US-based "first nationwide public service campaign in 1980" by a named entity for this exact issue wasn't definitively identified, organizations like Helen Keller Intl were actively working on implementing programs to prevent xerophthalmia due to vitamin A deficiency during this period, likely as part of broader public health or nutritional initiatives that may have included public awareness components. The impact of their work and the research from that era significantly contributed to the global understanding and efforts to combat this preventable cause of childhood blindness.\\
\end{promptbox}
    
\caption{Case study showing MLLM's shortcut reasoning behavior on a multi-hop query.}
\label{fig:case_study_MLLM}
\end{figure*}

\begin{figure*}[thb]
    \centering

    \begin{promptbox}{Query Reviewer Prompt}
    \textbf{System Prompt:}
    \textbf{Task} \\
    I have some QA data here, and you can observe that the questions can be divided into two categories:
    
    The category \#A: When you see this question alone without a given document, you are sure to find a unique document in a corpus to provide a unique answer. The question having some key words to help you locate the document from corpus. \\
    The category \#B: When you see this question alone without a given document, you will find hard to locate a document to give a deterministic answer for this question, because you will find multiple candidate documents in a corpus, which may lead to different answers for this question. The question do not have any special key words to help you locate the document from corpus.
    
    \textbf{Examples} \\
    The number mentioned on the right of the leftside margin? \#B \\
    What is the date mentioned in the second table? \#B \\
    What is the full form of PUF? \#A \\
    What is the number at the bottom of the page, in bold? \#B \\
    Who presented the results on cabin air quality study in commercial aircraft? \#A \\
    What is the name of the corporation? \#B \\
    Which part of Virginia is this letter sent from? \#B \\
    who were bothered by cigarette odors? \#A \\
    which cigarette would be better if offered on a thicker cigarette? \#A \\
    Cigarettes will be produced and submitted to O/C Panel for what purpose? \#A \\
    What is the heading of first table? \#B \\
    What is RIP-6 value for KOOL KS? \#A \\
    Which test is used to evaluate ART menthol levels that has been shipped? \#A \\
    How much percent had not noticed any difference in the odor of VSSS? \#A \\
    What is the cigarette code of RIP-6(W/O Filter) 21/4SE? \#A \\
    What mm Marlboro Menthol were subjectively smoked by the Richmond Panel? \#A \\
    What are the steps of Weft Preparation between Spinning bobbin and Weaving? \#A \\
    What level comes between Middle Managers and Non-managerial Employees? \#A \\
    What are the six parts of COLLABORATION MODEL of the organization where James has a role of leading the UK digital strategy? \#A
    \end{promptbox}
    
    \caption{Screening prompt for pilot study.}
    \label{fig:pilot-study-prompt}
\end{figure*}

\begin{figure*}[thb]
    \centering
    \begin{promptbox}{Visual Filter System Prompt}
    \textbf{System Prompt:} \\
    \textbf{Task:} \\
    You are a strict document image analyst for document-based RAG systems, 
    specializing in evaluating whether images can support the generation of high-quality Q\&A pairs.
    Carefully evaluate each image's content following these guidelines. \\[0.5em]
    
    \textbf{Core Objective:} \\
    Analyze the provided image and its surrounding page text to determine whether the content of the image can effectively support generating factually grounded QA pairs for document-based RAG systems. \\[0.5em]
    
    \textbf{Evaluation Criteria:}
    \begin{enumerate}[leftmargin=2em]
        \item \textbf{Essential Content Requirements:}
        \begin{itemize}
            \item The image must exhibit semantic cohesion with its surrounding text content.
            \item Visual elements should convey self-contained informational completeness.
            \item Demonstrate capacity to generate verifiable factual statements.
        \end{itemize}
        
        \item \textbf{Exclusion Imperatives:}
        Automatically reject images displaying:
        \begin{itemize}
            \item Decorative non-functionality.
            \item Semantic ambiguity preventing definitive interpretation.
            \item Information density below operational thresholds.
            \item Contextual detachment from document flow.
            \item Appendices, reference lists, or other images lacking specific meaningful information.
        \end{itemize}
    \end{enumerate}
    
    \textbf{Response Requirements:}
    \begin{itemize}[leftmargin=2em]
        \item Strictly respond with \texttt{"Yes"} or \texttt{"No"} only.
        \item \texttt{"Yes"} indicates the visual/textual content of the image can effectively support generating factually grounded QA pairs.
        \item \texttt{"No"} indicates the visual/textual content of the image cannot effectively support generating factually grounded QA pairs.
    \end{itemize}
    \end{promptbox}

    \caption{System prompt for evaluating visual content in document-Based RAG.}
    \label{fig:visual-filter-system-prompt}
\end{figure*}
\begin{figure*}[thb]
    \centering
    \begin{promptbox}{Singlehop QA Generation Prompt (Text)}
    Generate one QA pair based on the following guideline:
    
    \textbf{Question Requirements:}
    \begin{itemize}[leftmargin=2em]
        \item Create self-contained questions requiring no contextual knowledge from other pages.
        \item Focus on explicitly mentioned key entities, concepts, or processes.
        \item Avoid page-specific references (e.g., "in this section" or "as mentioned above").
        \item Include both factual questions (who/what/when) and explanatory questions (how/why).
    \end{itemize}

    \textbf{Answer Specifications:}
    \begin{itemize}[leftmargin=2em]
        \item Answers may be moderately summarized but must strictly adhere to source content.
        \item Prohibit any speculation or supplementation beyond original text.
    \end{itemize}

    \textbf{Format Rules:}
    \begin{itemize}[leftmargin=2em]
        \item Response must be in JSON format containing \texttt{"question"} and \texttt{"answer"} fields.
    \end{itemize}

    \textbf{Example Response:} \\
    \texttt{\{\\ "question": "What are the clinical diagnostic criteria for Parkinson's disease?", "answer": "Diagnosis requires bradykinesia combined with either resting tremor or rigidity, often presenting with asymmetric onset."\\ \}}
    \end{promptbox}

    \caption{Single-hop QA prompt template for generating self-contained question-answer pairs.}
    \label{fig:singlehop-qa-prompt}
\end{figure*}

\begin{figure*}[thb]
    \centering
    \begin{promptbox}{Singlehop QA Generation Prompt (Figure/Table)}
    Generate a QA pair based on the following guideline:

    \textbf{Question Requirements:}
    \begin{itemize}[leftmargin=2em]
        \item Formulate globally valid questions without image-dependent references.
              Expressions such as \texttt{"in the image/table"} or \texttt{"according to"} are prohibited.
        \item Focus on key visual elements: meaningful knowledge and insights rather than superficial visual elements.
        \item Prioritize questions that:
        \begin{itemize}
            \item Extract domain-specific knowledge.
            \item Identify key patterns and relationships.
            \item Explore underlying mechanisms or principles.
            \item Analyze trends and their implications.
            \item Examine cause-and-effect relationships.
        \end{itemize}
        \item Avoid trivial questions about:
        \begin{itemize}
            \item Simple counting of visual elements.
            \item Basic visual descriptions.
            \item Surface-level observations.
            \item Generic image properties.
        \end{itemize}
        \item Employ diverse question types and perspectives:
              \texttt{What / When / How / Where / Which}.
    \end{itemize}

    \textbf{Answer Specifications:}
    \begin{itemize}[leftmargin=2em]
        \item Answers must strictly derive from image content but not captions/context.
        \item Prohibit extrapolation beyond visually verifiable information.
        \item Focus on providing substantive, knowledge-rich responses.
    \end{itemize}

    \textbf{Format Rules:}
    \begin{itemize}[leftmargin=2em]
        \item Response must be in JSON format containing \texttt{"question"} and \texttt{"answer"} fields.
    \end{itemize}

    \textbf{Example Response:} \\
    \texttt{\{\\
      \ \ "question": "How does the introduction of carbon pricing mechanisms correlate with industrial emission reduction rates across different sectors in 2009?",\\
      \ \ "answer": "Manufacturing sectors showed a 30\% reduction in emissions after carbon pricing implementation, while energy sectors achieved a 45\% reduction, demonstrating stronger responsiveness to the policy."\\
    \}}
    \end{promptbox}

    \caption{Single-hop QA generation prompt for image-based question-answering.}
    \label{fig:singlehop-qa-image-prompt}
\end{figure*}
\begin{figure*}[thb]
    \centering
    \begin{promptbox}{Singlehop  Filter Prompt(Text)}
    You are a professional QA pair screening specialist with expertise in information quality assessment. 
    Your responsibility is to filter the QA pair for retrieval systems.
    Carefully analyze the QA pair and read the relevant context, 
    determine whether the question is worth keeping according to the following requirements. \\[0.5em]

    \textbf{Filter Requirements:}
    \begin{enumerate}[leftmargin=2em]
        \item \textbf{Question must be clear, self-contained, and explicitly reference entities/concepts.}
        \begin{itemize}
            \item \textbf{Most important: Reject questions containing} 
            \texttt{"according to the text"}, \texttt{"in the given information"}, 
            \texttt{"in the provided text"} \textbf{or similar phrases}.
            \item Reject vague or non-specific questions.
            \item Reject excessive inferences not mentioned in the original text.
            \item Reject queries based on appendices, reference lists, or images that do not contain specific meaningful information.
            \item Only retain questions about practical facts, data, processes, or concrete concepts.
        \end{itemize}

        \item \textbf{Answer must be fully supported by the provided text.}
        \begin{itemize}
            \item Reject answers not directly extractable from the text.
            \item Reject answers with factual errors or hallucinations.
        \end{itemize}
    \end{enumerate}

    \textbf{Output Format:}
    \begin{itemize}[leftmargin=2em]
        \item Respond with JSON format containing two keys:
        \begin{itemize}
            \item \texttt{"reason"}: Brief explanation for your decision (1--2 sentences).
            \item \texttt{"keep"}: \texttt{"Yes"} (meets ALL criteria, worth keeping) or \texttt{"No"} (fails ANY criterion).
        \end{itemize}
        \item Example: \\
        \texttt{\{\\
          \ \ "reason": "Question is clear and answer is well-supported by the text",\\
          \ \ "keep": "Yes"\\
        \}}
    \end{itemize}
    \end{promptbox}

    \caption{Single-hop text filter prompt for QA pair screening.}
    \label{fig:singlehop-text-filter-prompt}
\end{figure*}
\begin{figure*}[thb]
    \centering
    \begin{promptbox}{Singlehop Filter Prompt(Figure/Table)}
    You are a professional QA pair screening specialist with expertise in information quality assessment. 
    Your responsibility is to filter the QA pair for retrieval systems.
    Carefully analyze the QA pair, read the images and relevant context, 
    determine whether the question is worth keeping according to the following requirements. \\[0.5em]

    \textbf{Filter Requirements:}
    \begin{enumerate}[leftmargin=2em]
        \item \textbf{Question must be clear, self-contained, and explicitly reference entities/concepts.}
        \begin{itemize}
            \item \textbf{Most important: Reject questions containing} 
            \texttt{"according to the table"}, \texttt{"in the image"}, 
            \texttt{"in the given data"}, \texttt{"in the provided text"} \textbf{or similar phrases}.
            \item Reject vague or non-specific questions.
            \item Reject excessive inferences not mentioned in the original text.
            \item Reject queries based on appendices, reference lists, or images that do not contain specific meaningful information.
            \item Only retain questions about practical facts, data, processes, or concrete concepts.
        \end{itemize}

        \item \textbf{Answer must be fully supported by the provided text.}
        \begin{itemize}
            \item Reject answers not extractable from the image.
            \item Reject answers with factual errors or hallucinations.
        \end{itemize}
    \end{enumerate}

    \textbf{Output Format:}
    \begin{itemize}[leftmargin=2em]
        \item Respond with JSON format containing two keys:
        \begin{itemize}
            \item \texttt{"reason"}: Brief explanation for your decision (1--2 sentences).
            \item \texttt{"keep"}: \texttt{"Yes"} (meets ALL criteria, worth keeping) or \texttt{"No"} (fails ANY criterion).
        \end{itemize}
        \item Example: \\
        \texttt{\{\\
          \ \ "reason": "Question references specific visual elements and answer is supported by the image",\\
          \ \ "keep": "Yes"\\
        \}}
    \end{itemize}
    \end{promptbox}

    \caption{Single-hop image filter prompt for QA pair screening.}
    \label{fig:singlehop-image-filter-prompt}
\end{figure*}
\begin{figure*}[thb]
    \centering
    \begin{promptbox}{Table-to-Text Caption Prompt}
    \textbf{Role:} \\
    You are a data analyst specializing in precise verbalization of structured data. \\[0.5em]

    \textbf{Task:} \\
    Convert tabular data from the document's contextual documents into natural language descriptions. \\[0.5em]

    \textbf{Core Requirements:}
    \begin{itemize}[leftmargin=2em]
        \item Use table entities/objects as direct grammatical subjects without mentioning the table structure.
        \item \textbf{Do not} begin the description with phrases such as 
              \texttt{"the survey"}, \texttt{"the table"}, or \texttt{"the figure"}.
        \item The description must be clear and understandable even when taken out of the original context, 
              ensuring it can clearly refer to and express the intended meaning and be unambiguously answered.
        \item Exhaustively describe every data cell using original values and labels.
        \item Maintain absolute objectivity – no analysis, interpretations, or subjective terms.
        \item Form cohesive paragraphs using transitional phrases (no bullet points/list formats).
        \item Embed specific context from source documents into descriptions.
        \item If the table contains footnotes or annotations, include their explanations.
    \end{itemize}
    \end{promptbox}

    \caption{Table-to-text caption prompt for converting tabular data into descriptive text.}
    \label{fig:table-to-text-caption-prompt}
\end{figure*}
\begin{figure*}[thb]
    \centering
    \begin{promptbox}{Figure-to-Text Caption Prompt}
    \textbf{Role:} \\
    You are a visual analyst specialized in exhaustive objective description of visual content. \\[0.5em]

    \textbf{Task:} \\
    Generate comprehensive descriptions of images strictly based on their pictorial content and document context. \\[0.5em]

    \textbf{Core Requirements:}
    \begin{itemize}[leftmargin=2em]
        \item Use depicted entities/objects as grammatical subjects (prohibited: 
              \texttt{"The bar chart shows..."}).
        \item \textbf{Do not} begin the description with 
              \texttt{"the survey"}, \texttt{"the table"}, or \texttt{"the figure"}.
        \item The description must be clear and understandable even when taken out of the original context, 
              ensuring it can clearly refer to and express the intended meaning and be unambiguously answered.
        \item Describe \textbf{all} visual elements:
        \begin{itemize}
            \item For infographics: Every data point, axis labels, trend lines, flow directions, and legend entries.
            \item For objects/people: Physical attributes, spatial relationships, and observable actions.
        \end{itemize}
        \item Maintain objectivity:
        \begin{itemize}
            \item No subjective terms.
            \item No analytical conclusions (e.g., \texttt{"This suggests..."}).
            \item No contextual assumptions beyond provided documentation.
        \end{itemize}
        \item Preserve data and information integrity.
        \item \textbf{Form cohesive paragraphs using transitional phrases} (no bullet points/list/markdown formats).
        \item Embed specific context from source documents into descriptions.
    \end{itemize}
    \end{promptbox}

    \caption{Figure-to-text caption prompt for objective image description.}
    \label{fig:figure-to-text-caption-prompt}
\end{figure*}
\begin{figure*}[thb]
    \centering
    \begin{promptbox}{Relationship Evaluation and Selection Prompt}
    You are an expert in knowledge graph reasoning. Your task is to evaluate relationship 
    candidates and select the best one for constructing an unambiguous reasoning question. \\[0.5em]

    The ideal relationship should uniquely identify the target entity. 
    When forming a question like \texttt{"What entity [relation] with [current entity]?"}, 
    the answer should be specific enough that only one reasonable entity fits. 
    It is \textbf{strictly forbidden} to select vague relationships such as 
    \texttt{"is related to"}.
    \end{promptbox}

    \caption{Prompt for evaluating and selecting knowledge graph relationships.}
    \label{fig:relationship-evaluation-selection-prompt}
\end{figure*}
\begin{figure*}[thb]
    \centering
    \begin{promptbox}{Relationship Selection Prompt}
    Given the current entity \texttt{\{current\_node\}}, I have the following candidate entities 
    and their relations to the current entity:

    \texttt{\{candidates\_json\}}

    Please evaluate each relationship and select the \textbf{one} that would create the most unambiguous
    and specific reasoning question. The chosen relationship should make it possible to uniquely
    identify the target entity when given the current entity and the relationship. \\[0.5em]

    \textbf{Return your response as a JSON object with:}
    \begin{enumerate}[leftmargin=2em]
        \item \texttt{"reasoning"}: Brief explanation of why this relationship is the most specific/unique.
        \item \texttt{"selected\_index"}: The index (0-based) of the chosen candidate.
    \end{enumerate}

    \textbf{Example response format:} \\
    \texttt{\{\\
      \ \ "reasoning": "This relationship 'is the inventor of' creates the most unique connection...",\\
      \ \ "selected\_index": 2\\
    \}}
    \end{promptbox}

    \caption{Prompt for selecting the most unambiguous knowledge graph relationship.}
    \label{fig:relationship-selection-prompt}
\end{figure*}
\begin{figure*}[thb]
    \centering
    \begin{promptbox}{Step Question Generation Prompts}
    \textbf{General Prompt:} \\
    Generate a simple question (Q) and answer (A) pair about the relationship between two entities. 
    Given an entity and a relation, ask for the entity at the other end. 
    The answer should be the specific entity name provided. 
    Return the response as a JSON object with keys \texttt{"question"} and \texttt{"answer"}. \\
    \textbf{Note:} If the answer is in all uppercase letters, you must convert it to the appropriate case. \\[0.5em]

    \textbf{User-Specific Prompt:} \\
    Given Entity \texttt{\{current\_node\_id\}} and the relationship \texttt{\{relation\_text\}}, 
    generate a question (Q) that asks for the entity connected by this relationship. 
    The answer (A) is \texttt{\{next\_node\_id\}}. \\
    \textbf{Note:} If the answer is in all uppercase letters, you must convert it to the appropriate case. \\
    Return the response as a JSON object with keys \texttt{"question"} and \texttt{"answer"}.
    \end{promptbox}

    \caption{Step question generation prompts for entity-Relationship QA pairs.}
    \label{fig:step-qa-generation-prompts}
\end{figure*}
\begin{figure*}[thb]
    \centering
    \begin{promptbox}{Multi-hop Question Chaining Prompts}
    \textbf{General Prompt:} \\
    Combine two questions to form a natural-sounding multi-hop reasoning question. \\[0.5em]

    \textbf{Guidelines:}
    \begin{enumerate}[leftmargin=2em]
        \item Analyze both questions and identify exactly how the entity appears in Q2.
        \item Consider different ways to phrase the combined question that sound natural.
        \item Think about what phrasing would be most clear to a human reader.
        \item Reason about whether any ambiguity might be introduced by the combination.
    \end{enumerate}

    \textbf{Goal:} \\
    Seamlessly integrate the first question into the second question by replacing a specific entity reference. 
    The combined question should:
    \begin{itemize}[leftmargin=2em]
        \item Be grammatically correct and flow naturally.
        \item \textbf{Avoid awkward phrases} like \texttt{"the entity that..."} or \texttt{"the thing which..."}.
        \item Maintain the original meaning and logical connection between questions.
        \item Sound like a question a human would ask, not an artificial construction.
        \item Accurately preserve the reasoning chain between the questions.
    \end{itemize}

    After your analysis, provide the final combined question in \textbf{JSON format}. \\[1em]

    \textbf{User Prompt:} \\
    Combine the following questions: \\
    Question 1 (Q1): \texttt{\{previous\_cumulative\_q\}} \\
    Question 2 (Q2): \texttt{\{new\_step\_q\}} \\
    Entity to replace: \texttt{\{entity\_to\_replace\}} \\[0.5em]

    First, explain your reasoning: analyze how you will approach combining these questions naturally. 
    Think about how to best replace \texttt{\{entity\_to\_replace\}} with Q1 in a way that reads fluently. \\

    Then, provide your final combined question as a JSON object with key \texttt{"chained\_question"}. \\[0.5em]

    \textbf{Examples:}
    \begin{itemize}[leftmargin=2em]
        \item Example 1: \\
        Q1: \texttt{"What is the capital of France?"} \\
        Q2: \texttt{"What river flows through Paris?"} \\
        Combined: \texttt{"What river flows through the capital of France?"}
        \item Example 2: \\
        Q1: \texttt{"Who directed Pulp Fiction?"} \\
        Q2: \texttt{"What other movies did Quentin Tarantino make?"} \\
        Combined: \texttt{"What other movies did the director of Pulp Fiction make?"}
    \end{itemize}

    \textbf{JSON format example:} \\
    \texttt{\{\\
      \ \ "chained\_question": "Your combined question here"\\
    \}}
    \end{promptbox}

    \caption{Multi-hop question chaining prompts for combining reasoning steps.}
    \label{fig:multi-hop-question-chaining}
\end{figure*}
\begin{figure*}[thb]
    \centering
    \begin{promptbox}{Multi-hop QA Filter Prompt}
    {\tiny
    You are a strict expert in knowledge graph reasoning and natural language question generation quality assessment. 
    Your task is to rigorously evaluate a provided multi-hop reasoning question and its underlying reasoning steps based on 
    a set of strict quality criteria. You must analyze the entire reasoning path, from the initial step to the final question and answer, 
    to determine if the question is high-quality, unambiguous, logically sound, and meaningful. \\[0.5em]

    Evaluate the following multi-hop reasoning question and its construction process based on the criteria provided below. \\[0.5em]

    \textbf{Question Data:} \\
    Initial Question: \texttt{\{initial\_question\}} \\
    Initial Answer (Entity ID): \texttt{\{initial\_answer\}} \\[0.5em]
    Final Question: \texttt{\{final\_question\}} \\
    Final Answer (Entity ID): \texttt{\{final\_answer\}} \\[0.5em]
    \textbf{Reasoning Steps (Logical Flow and Chaining):} \\
    \texttt{\{steps\_description\}} \\[0.8em]

    \textbf{Evaluation Criteria:}
    \begin{enumerate}[leftmargin=2em]
        \item \textbf{Final Question Clarity and Context Independence:}
        \begin{itemize}
            \item The \texttt{Final Question} must be clearly, fluently, and naturally phrased and must be understandable and answerable.
            \item The \texttt{Final Answer} entity name must \textbf{NOT} be present within the \texttt{Final Question} text itself.
            \item The \texttt{Final Question} should be specific enough to uniquely identify the \texttt{Final Answer} given the preceding context.
            \item The \texttt{Final Question} and all step questions must avoid artificial phrasing like 
                  \texttt{"the entity that..."}, \texttt{"the thing which..."}, etc., and sound like natural human questions.
        \end{itemize}

        \item \textbf{Necessity of Reasoning Steps:}
        \begin{itemize}
            \item Every intermediate step described in the \texttt{Reasoning Steps} must be logically essential to derive the \texttt{Final Answer}.
            \item For each \texttt{step question}, the answer should correspond exactly to the question and not be random or irrelevant.
            \item Removing any intermediate step should break the logical chain required to answer the \texttt{Final Question}.
        \end{itemize}

        \item \textbf{Rigor and Uniqueness of Steps:}
        \begin{itemize}
            \item Each step (\texttt{current\_node} → \texttt{next\_node} via \texttt{relation\_text}) must be specific enough to imply a unique \texttt{answer} (\texttt{next\_node}) 
                  within the context of a typical broad knowledge base \textbf{(MOST IMPORTANT)}.
            \item The answer entity for any step must \textbf{NOT} appear directly within the \texttt{Step Question}.
            \item The cumulative questions should correctly integrate each step question.
        \end{itemize}

        \item \textbf{Significance:}
        \begin{itemize}
            \item The \texttt{Final Question} must address a meaningful query about the entities and their relationships.
            \item It should not be trivial, overly generic, or based on obscure, impractical connections.
        \end{itemize}
    \end{enumerate}

    Based on your evaluation, first provide a brief textual explanation (\texttt{reason}) summarizing your evaluation and conclusion (why it passed or failed). \\[0.5em]

    Then, provide your final decision as a JSON object with two keys:
    \begin{itemize}
        \item \texttt{"reason"}: (string) Your explanation as described above.
        \item \texttt{"keep"}: (string) Must be either \texttt{"yes"} or \texttt{"no"}.
    \end{itemize}

    Return \textbf{ONLY the JSON object} after your explanation.
    }
    \end{promptbox}

    \caption{Multi-hop QA filter prompt for evaluating reasoning questions.}
    \label{fig:multihop-qa-filter-prompt}
\end{figure*}
\begin{figure*}[thb]
    \centering
    \begin{promptbox}{Page Ground Truth (GT) Prompt}
    You are a professional document analysis expert tasked with determining page relevance. 
    Follow these guidelines precisely: \\[0.5em]

    \textbf{Task Definition:} \\
    Analyze the provided document page along with the question-answer pair to determine if the page contains relevant information. \\[0.5em]

    \textbf{Decision Criteria:}
    \begin{itemize}[leftmargin=2em]
        \item Respond with \texttt{"Yes"} \textbf{ONLY} if \textbf{ALL} of these conditions are met:
        \begin{itemize}
            \item The page's information is \textbf{ESSENTIAL} for understanding the query and answer.
        \end{itemize}

        \item Respond with \texttt{"No"} if:
        \begin{itemize}
            \item The page contains no information related to the question.
            \item The information is only tangentially related and provides minimal value.
            \item The page contains partial information that requires significant inference or external knowledge to answer the question.
        \end{itemize}
    \end{itemize}

    \textbf{Format Instructions:} \\
    You must respond \textbf{ONLY} with \texttt{"Yes"} or \texttt{"No"} — no explanations or additional text.
    \end{promptbox}

    \caption{Page ground truth prompt for document relevance assessment.}
    \label{fig:page-gt-prompt}
\end{figure*}

\begin{figure*}[thb]
    \centering
    \begin{promptbox}{Question Refinement  Prompt}
    You are an expert specializing in query analysis and refinement for a multimodal Retrieval-Augmented Generation (RAG) system. Your primary function is to disambiguate user questions by making them more specific.\\

    \textbf{Task Definition:} \\
   Your objective is to refine an ambiguous Original Question by leveraging a designated Ground Truth Image and contrasting it with several irrelevant distractor images. \\[0.5em]

    \textbf{Instructions:}
    
    \begin{itemize}[leftmargin=2em]
        \item You will be given a question, a single Ground Truth Image (which will always be the first image provided), and one or more distractor images.

        \item Your task is to analyze the Original Question and compare the Ground Truth Image against the distractor images.By contrasting them, identify the key, distinguishing detail within the Ground Truth Image that makes it the unique and correct answer to the question. This detail should be absent or different in the distractor images.

        \item Finally, rewrite the Original Question to create a new, more precise Refined Question that seamlessly incorporates this key detail, making the question unambiguous.
        \\[0.5em]
    \end{itemize}

    \textbf{Format Instructions:} \\
    Your response MUST be a valid JSON object with exactly two keys::
    \begin{itemize}[leftmargin=2em]
        \item reason:  A string containing your reasoning process.

        \item question: Your refined question.
    \end{itemize}
    \end{promptbox}

    \caption{Refinement  Prompt for Ambiguous Question}
    \label{fig:Question-Refinement-Prompt}
\end{figure*}

\begin{figure*}[thb]
    \centering
    \begin{promptbox}{Answer Generation Prompt(Oracle)}
    You are a professional document analyst\\

    \textbf{Task Definition:} \\
    Your primary task is to thoroughly analyze the user-provided image(s) to answer their question. Your answer must be supported by the visual evidence and details found within the image(s) \\[0.5em]

    \textbf{Instruction:}
    \begin{itemize}[leftmargin=2em]
        \item First, formulate a brief reasoning process. This should explain how you derived your answer from the visual evidence and any contextual knowledge you applied.

        \item Second, provide a concise and direct final answer in \{language\}.\\[0.5em]
    \end{itemize}

    \textbf{Format Instructions:} \\
    Your entire response MUST be a single, valid JSON object. This JSON object must contain exactly two keys:
    \begin{itemize}[leftmargin=2em]
        \item reason: A string containing your English reasoning process.

        \item answer:A string containing your final answer in the requested language.
    \end{itemize}
    \end{promptbox}

    \caption{Answer generation prompt for oracle experiments.}
    \label{fig:answer generation prompt oracle}
\end{figure*}

\begin{figure*}[thb]
    \centering
    \begin{promptbox}{Answer Evaluation Prompt}
    You are a fair and objective grader. Your judgment should be based on a balanced assessment.\\

    \textbf{Task Definition:} \\
    Your primary task is to evaluate the \texttt{"Generated Answer"} by comparing it against the \texttt{"Ground Truth Answer"}, taking into account the original Question. Based on this evaluation, you will assign a single integer score from 1 to 10. \\[0.5em]

    \textbf{Scoring Rubric:}
    You must adhere to the following 1-10 scale.
    \begin{itemize}[leftmargin=2em]
        \item Score 1-3 (Poor): The answer is largely incorrect, irrelevant, contains significant inaccuracies or hallucinations, or demonstrates a fundamental misunderstanding of the question.

        \item Score 4-6 (Acceptable): The answer is partially correct but either misses important information, is somewhat vague, or contains minor inaccuracies. It shows some understanding but is not comprehensive.

        \item Score 7-8 (Good): The answer is correct and aligns well with the ground truth. It covers most key aspects but might lack a few minor details, could be slightly less concise, or have some minor phrasing improvements.

        \item Score 9-10 (Excellent): The answer is fully correct, complete, and concise. It accurately captures all essential aspects of the ground truth and is well-articulated.
        \\[0.5em]
    \end{itemize}

    \textbf{Format Instructions:} \\
    Your response MUST be a valid JSON object with exactly two keys::
    \begin{itemize}[leftmargin=2em]
        \item reason: reason: A brief, one-sentence justification for your score .

        \item score: An integer from 1 to 10.
    \end{itemize}
    \end{promptbox}

    \caption{Answer evaluation prompt for main experiments.}
    \label{fig:qa-judge-prompt}
\end{figure*}

\bibliographystyle{iclr2025}

\end{document}